\PassOptionsToPackage{numbers,sort&compress}{natbib}
\documentclass{article}

\usepackage[preprint]{neurips_2026}

\usepackage{amsmath, amssymb, amsthm}
\usepackage{booktabs}
\usepackage{graphicx}
\usepackage{hyperref}
\usepackage{url}
\usepackage{microtype}
\usepackage{xcolor}
\usepackage{multirow}
\usepackage{subcaption}
\usepackage{algorithm}
\usepackage{fancyvrb}
\usepackage{algpseudocode}
\usepackage{wrapfig}
\usepackage{colortbl}
\usepackage{enumitem}
\usepackage{listing}

\newcommand{\method}{\textsc{Gated Recurrent Transformer}}  
\newcommand{\methodshort}{\textsc{GRT}}                     

\newcommand{\repo}{\url{https://github.com/Amr-Hegazy1/gated-recurrent-transformer}}

\newcommand{\prelude}{n_{\text{pre}}}
\newcommand{\shared}{n_{\text{shared}}}
\newcommand{\coda}{n_{\text{coda}}}
\newcommand{\nrec}{n_{\text{rec}}}
\newcommand{\recmax}{R}           
\newcommand{\recinf}{R^*}         

\newcommand{\Bshared}{\mathcal{B}_{\text{shared}}}

\newcommand{\hstate}{\mathbf{h}}

\newcommand{\gate}{\mathbf{g}}

\newcommand{\flops}{\text{FLOPs}}

\newcommand{\isoflop}{\textit{iso\textsc{FLOPs}}}
\newcommand{\isoparam}{\textit{iso\textsc{Params}}}

\newcommand{\reals}{\mathbb{R}}
\newcommand{\sigmoid}{\sigma}

\newcommand{\secref}[1]{Section~\ref{#1}}
\newcommand{\tabref}[1]{Table~\ref{#1}}
\newcommand{\figref}[1]{Figure~\ref{#1}}
\newcommand{\eqrefx}[1]{Eq.~(\ref{#1})}
\newcommand{\listref}[1]{Listing~\ref{#1}}

\definecolor{beatcol}{RGB}{0,128,85}

\usepackage[most]{tcolorbox}
\newtcolorbox{takeaway}[1][]{
  colback=blue!5!white,
  colframe=blue!40!black,
  fonttitle=\bfseries\small,
  title={Takeaway},
  boxrule=0.4pt,
  left=4pt, right=4pt, top=3pt, bottom=3pt,
  #1
}

\definecolor{lst@kw}{rgb}{0.13,0.29,0.53}
\definecolor{lst@cm}{rgb}{0.20,0.45,0.20}
\newcommand{\kw}[1]{\textcolor{lst@kw}{\textbf{#1}}}
\newcommand{\cmt}[1]{\textcolor{lst@cm}{\textit{#1}}}

\title{Gated Recurrent Transformers: Expressive Depth through Recurrent Modulation}

\author{
  Amr Hegazy \\
  The German University in Cairo \\
  \texttt{amr.hazem@student.guc.edu.eg} \\
  \And
  Amr Alanwar \\
  Technical University of Munich \\
  \texttt{alanwar@tum.de} \\
  \And
  Mostafa Elhoushi \\
  Cerebras Systems Inc. \\
  \texttt{mostafa.elhoushi@cerebras.net} \\
}

\begin{document}

\maketitle

\begin{abstract}

Scaling transformer language models creates an inherent tension between expressivity and memory efficiency. 
While unique weights across layers preserve functional specialization—from input-grounding to abstract refinement—they incur a substantial memory footprint. 
Conversely, standard depth-sharing enforces uniform transformations that collapse representational diversity and degrade modeling quality.
We introduce the \method{} (\methodshort{}), a recurrent depth transformer where fixed-depth prelude and coda blocks bracket a single shared core iterated $R$ times.
Inspired by gated recurrent neural networks, we employ a lightweight projection and an elementwise update gate---conditioned on the hidden state, the fixed prelude output, and noise resampled at every step---to modulate the recurrent update. This allows the model to specialize the input to the same few layers across recurrences, rather than requiring many unique layers to achieve functional diversity. 
Under an \isoflop{} constraint,
a 3-layer \method{} matches the accuracy of a 12-layer GPT-2 Small baseline
with similar training and inference FLOPs, and leads MoR and heavy-tail depth
sampling in all nine scale-by-budget cells; at medium and large scale
it approaches dense quality at the standard token budget and overtakes it at
medium scale once that budget is doubled.
Under an \isoparam{} constraint,
deeper recurrence achieves a \textbf{2.76} validation loss
versus \textbf{2.84} for a non-recurrent counterpart at matched parameter and data budget.
Our results demonstrate that adaptive depth reuse is a principled
strategy for trading parameters for quality: at large scale, $62\%$ fewer
parameters and $59\%$ less peak decoding memory for a $10\%$ increase in
compiled generation latency. 
Code is available at \repo.
\end{abstract}

\section{Introduction}
\label{sec:intro}

Scaling transformer language models has delivered remarkable gains, yet it necessitates a simultaneous increase in parameter count and training compute~\citep{kaplan2020scalinglawsneurallanguage}. In standard architectures, depth and parameters are rigidly coupled: every additional layer introduces a fresh set of weights, creating a memory bottleneck that constrains effective depth under fixed hardware budgets. 

Beyond memory efficiency, however, there is a deeper computational motivation rooted in the nature of intelligence itself: Turing's foundational insight~\citep{turing1936computable} is that a \emph{finite} set of states and symbols, applied \emph{iteratively}, is sufficient to compute anything computable — suggesting that the power of a reasoning system lies not in its breadth of parameters, but in its depth of iteration. This principle finds a striking modern echo in the test-time compute paradigm~\citep{snell2025scaling}, where models allocate more compute at inference to improve reasoning~\citep{openai2024o1, deepseekai2025r1}. While existing approaches realize this compute through chain-of-thought steps in the output space—consuming sequence length in the process~\citep{feng2023towards}—recurrent depth offers a complementary realization: by iterating a shared transformation over the input representation, the model ``thinks'' more deeply entirely within its hidden states without generating extra tokens or storing extra parameters.

Weight sharing is the natural response to memory-depth coupling, allowing effective depth to grow without inflating parameter count~\citep{lan2020albert, universaltransformers, huggin2024, ouro}. Prior work has utilized recurrent depth either as an efficiency tool to match quality at reduced parameter counts~\citep{jeddi2026loopformer, bae2025relaxed} or as a performance tool to improve accuracy at matched parameter counts~\citep{huggin2024, ouro}. However, a unified treatment of these perspectives is missing, and a deeper architectural tension remains: standard weight sharing forces an identical transformation on an ever-evolving representation. Because the hidden state after the first recurrence differs fundamentally from the eighth, a static transformation collapses the functional diversity that makes depth valuable~\citep{vld2025}. This raises the question: Can a single fixed transformation serve meaningfully across such diverse representational stages, or does it inevitably collapse the functional diversity that makes depth valuable in the first place~\citep{vld2025}. Such questions motivate the design of \method{}.

%

To resolve this tension, \method{} introduces a per-element learned gate that conditions on the current hidden state, a fixed \emph{prelude} representation of the original input~\citep{huggin2024}, and stochastic noise. This constructs a distinct, context-grounded input at every recurrence, allowing a single weight tensor to behave as multiple specialized layers. Gates are initialized such that training starts with the residual stream passing through nearly unchanged through all recurrences. As training progresses, gating emerges gradually as the shared block learns to selectively refine representations. The shared transformation is thus fixed in weights but dynamic in behavior — the model thinks differently at each pass despite reusing the same parameters. Combined with depth sampling during training~\citep{huggin2024}, this architecture enables a continuous compute-quality tradeoff at inference and facilitates evaluation across both \isoflop{} and \isoparam{} regimes.

We summarize our contributions as follows:
\begin{itemize}[nosep]
    \item \textbf{Architecture}: We propose \method{}, which uses per-element gating and stochastic perturbations to prevent representational collapse, allowing a shared core to behave distinctly across recurrence steps.
    \item \textbf{Unified Evaluation}: We evaluate \method{} under both \isoflop{} and \isoparam{} regimes at three scales. Under \isoflop{} it leads MoR and heavy-tail depth sampling in all nine scale-by-budget cells and matches or leads RRT and Ouro at every scale at the standard budget, matches the dense GPT-2 baseline at small scale while using \textbf{64\%} fewer parameters, and closes on it at medium and large scale as the token budget grows. Under \isoparam{} it improves validation loss at matched parameter count, at a higher inference-FLOP cost.
    \item \textbf{Emergent Early Exit}: Without auxiliary losses, \method{} naturally supports inference-time early exiting, retaining $\sim$\textbf{92\%} accuracy when running only half of its recurrence steps.
\end{itemize}

\begin{figure}[t]
  \centering
  \includegraphics[width=\linewidth]{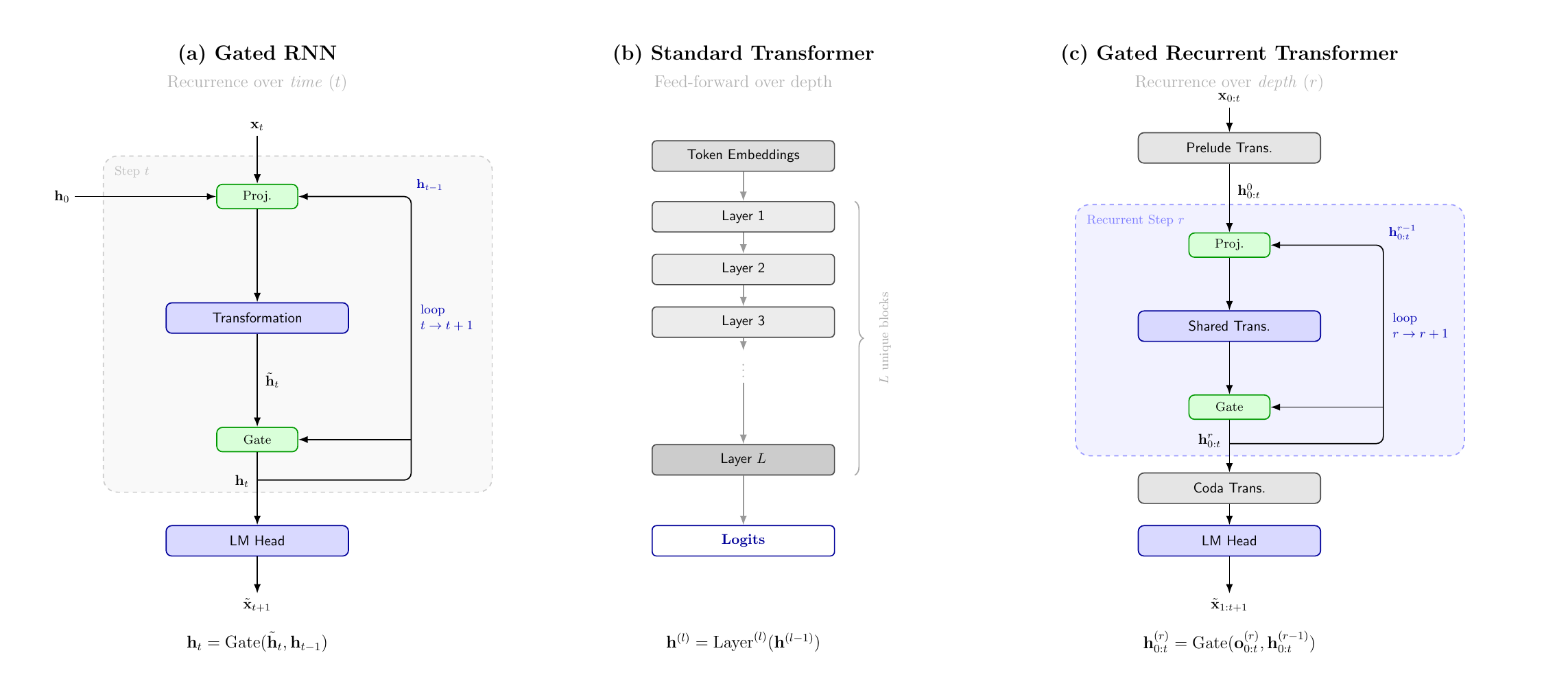}
  \caption{
    \textbf{\method{} overview.}
    (a)~A Gated Recurrent Unit (GRU) reuses the same weight matrices at every sequence
    position---recurrence over the \emph{sequence} dimension.
    (b)~A standard transformer stacks $L$ independent layers, each with
    unique weights, applied once to the full sequence.
    (c)~\method{} introduces recurrence over the \emph{depth} dimension, applying each recurrence to the full sequence 
  }
  \label{fig:teaser}
\end{figure}

\section{Related Work}
\label{sec:related}

The concept of shared weights across repeated computational steps traces back to the seminal 1986 backpropagation paper of  Rumelhart, Hinton, and Williams\citep{rumelhart1986learning}, who briefly proposed at the end of their paper the synchronous iterative net: a network where each iteration corresponds to a layer with tied weights. \citep{Werbos1990} extended this to temporal sequence modeling via backpropagation through time (BPTT), inspiring recurrent architectures — RNNs~\citep{ElmanRNN}, LSTMs~\citep{LSTM}, and GRUs~\citep{GRU} — that apply shared weights along the sequence dimension, processing inputs one timestep at a time. In contrast, this work applies recurrent depth across the full input, iterating a shared-weight transformation along network depth to progressively refine the input representation.

In the era of transformers, one of the earliest papers on recurrent depth was ALBERT~\citep{lan2020albert}, which proposed a BERT~\citep{bert} model with all transformer layers sharing the same weights. Universal Transformers~\citep{universaltransformers} further added a per-token adaptive halting mechanism. However, both methods shared the \emph{entire} layer stack, forcing every layer to apply an identical transformation regardless of the representational stage of the input. \citep{huggin2024} relaxed this constraint by assigning distinct roles to prelude, shared core, and coda layers. Similarly, \citet{koishekenov2025encodethinkdecodescaling} identified early, middle, and late layers of pretrained LLMs as encoding, reasoning, and decoding stages respectively, and proposed recurring only the middle layers while keeping the rest fixed. Our \method{} follows this principle of selectively recurring middle layers.

Prior work has shown that parameters in feed-forward network (FFN) layers primarily store factual knowledge \citep{geva-etal-2021-transformer}, while computational depth drives reasoning ability \citep{zhu2025parametersexploringvirtuallogic}, motivating recurrence to improve reasoning. We organize prior work on recurrent depth transformers along two evaluation regimes. In the \isoparam{} regime — increasing FLOPs by recurring layers while keeping parameter count fixed, trading compute for improved accuracy — \citep{huggin2024,ouro,koishekenov2025encodethinkdecodescaling} demonstrate consistent quality gains. In the \isoflop{} regime — reducing parameter count by recurring layers while matching training and inference compute, trading memory for efficiency — \citep{jeddi2026loopformer} and \citep{bae2025relaxed} show that depth reuse can achieve comparable quality at a smaller memory footprint. Our work spans both regimes, with \isoflop{} as our primary contribution and complementary \isoparam{} results reported in Section~\ref{sec:experiments}.

A related line of work adapts the amount of computation dynamically rather than fixing it uniformly across tokens or layers. Adaptive Computation Time (ACT)~\citep{graves2017adaptivecomputationtimerecurrent} halts computation per-position based on learned termination signals, allowing different tokens to consume different amounts of compute. Mixture-of-Depths~\citep{raposo2024mixtureofdepthsdynamicallyallocatingcompute} takes a routing approach, dynamically assigning tokens to different subsets of layers rather than processing all tokens through all layers. More recently, Mixture-of-Recursions~\citep{bae2026mixtureofrecursions} extends this idea to the recurrent setting, routing subsets of tokens to different numbers of recursion steps. While all three methods adapt computation at the token level — through halting, layer routing, or recursion routing — they require dedicated termination signals or routing mechanisms. In contrast, \method{} uses a simpler gating mechanism that modulates how much of the shared-block update is absorbed at each recurrence step, without any routing or halting logic.


Weight-tied iteration toward a fixed point is the defining feature of deep equilibrium models~\citep{bai2019deq}, which solve for the equilibrium directly and differentiate through it with implicit gradients, avoiding the memory cost of storing an unrolled trajectory. \method{} shares the weight-tied update but not the equilibrium objective: we unroll a fixed number of discrete steps and backpropagate through them, and depth sampling trains each step to serve as an exit rather than as an approach to a single limit.

Scaling laws~\citep{kaplan2020scalinglawsneurallanguage, chinchilla} have established how model size and data jointly determine performance, with Pythia~\citep{pythia} providing controlled comparisons across scales. Notably, \citet{kaplan2020scalinglawsneurallanguage} directly evaluated parameter-sharing transformers and observed that recurrent models perform better at matched parameter count (\isoparam{}) but worse at matched compute (\isoflop{}). The interplay between depth and width has received dedicated attention: \citep{yehudai2026depthwidthtradeoffsalgorithmicreasoning} theoretically shows that increasing width can compensate for reduced depth in algorithmic reasoning tasks, while \citep{mcleish2026gemstones} empirically demonstrates that scaling law prescriptions are sensitive to depth-width ratio. In our \method, we propose a novel recurrent depth architecture that aims to improve the scaling laws of loss versus parameter count and loss versus training FLOPs, demonstrating consistent gains under both \isoflop{} and \isoparam{} regimes as model size and training data scale.

\section{Recurrent Depth Reuse for Language Modeling}
\label{sec:method}

\subsection{Preliminaries}

Let $\mathbf{X} = (x_1, \ldots, x_T)$ be a token sequence of length $T$ drawn from vocabulary $\mathcal{V}$.
A standard autoregressive transformer with $L$ blocks defines a chain of residual updates:
$\hstate^{(\ell)} = \hstate^{(\ell-1)} + \mathrm{Block}_\ell(\hstate^{(\ell-1)})$ for $\ell = 1, \ldots, L$, 
where $d$ is the embedding dimension and $\hstate^{(0)} \in \reals^{T \times d}$ are initial token embeddings. 
Each block comprises pre-norm multi-head self-attention and a token-wise MLP. 
The objective is to minimize the negative log-likelihood, $\mathcal{L} = -\sum_t \log p(x_t \mid x_{<t})$. 
Standard transformers couple depth and parameters rigidly, as every additional layer adds a fresh set of weights. The total unique parameter count scales as $\Theta(L \cdot d^2)$.

\subsection{The Prelude--Shared--Coda Architecture}
\label{subsec:architecture}

\method{} follows~\citep{huggin2024} in partitioning transformer blocks into three sets, denoted by the shorthand 
$\prelude\texttt{+}\nrec\!\times\!\recmax\texttt{+}\coda$. 
We have observed that this separation of fixed context
encoders from a shared recurrent core yields more stable training than
uniform weight sharing across all layers.

\begin{itemize}[nosep]
    \item \textbf{Prelude ($\prelude$ blocks):} Applied once to produce a stable, context-aware conditioning signal $\hstate^{(\mathrm{pre})} = \mathrm{Block}_{\prelude}\circ\cdots\circ\mathrm{Block}_1(\hstate^{(0)})$.
    \item \textbf{Shared core ($\nrec$ blocks):} Applied recurrently $R$ times. The first step is initialized with stochastic noise $\epsilon_x$ conditioned on the prelude output. In subsequent steps, the core iteratively processes the previous hidden state $\hstate^{(r-1)}$ alongside the fixed prelude anchor: $\hstate^{(R)} = \mathrm{Block}_{\shared}\circ\cdots\circ\mathrm{Block}_1[\epsilon_x,\,\hstate^{(\mathrm{pre})}]$. We provide more details on the operation in the subsequent section.
    \item \textbf{Coda ($\coda$ blocks):} Receives the final refined state $\hstate^{(R)}$ and projects to output final hidden state $\hstate^{(\mathrm{coda})} = \mathrm{Block}_{\coda}\circ\cdots\circ\mathrm{Block}_1(\hstate^{(R)})$ that are then fed into the language model to produce logits.
\end{itemize}

\begin{figure}[t]
  \centering
  \includegraphics[width=\linewidth]{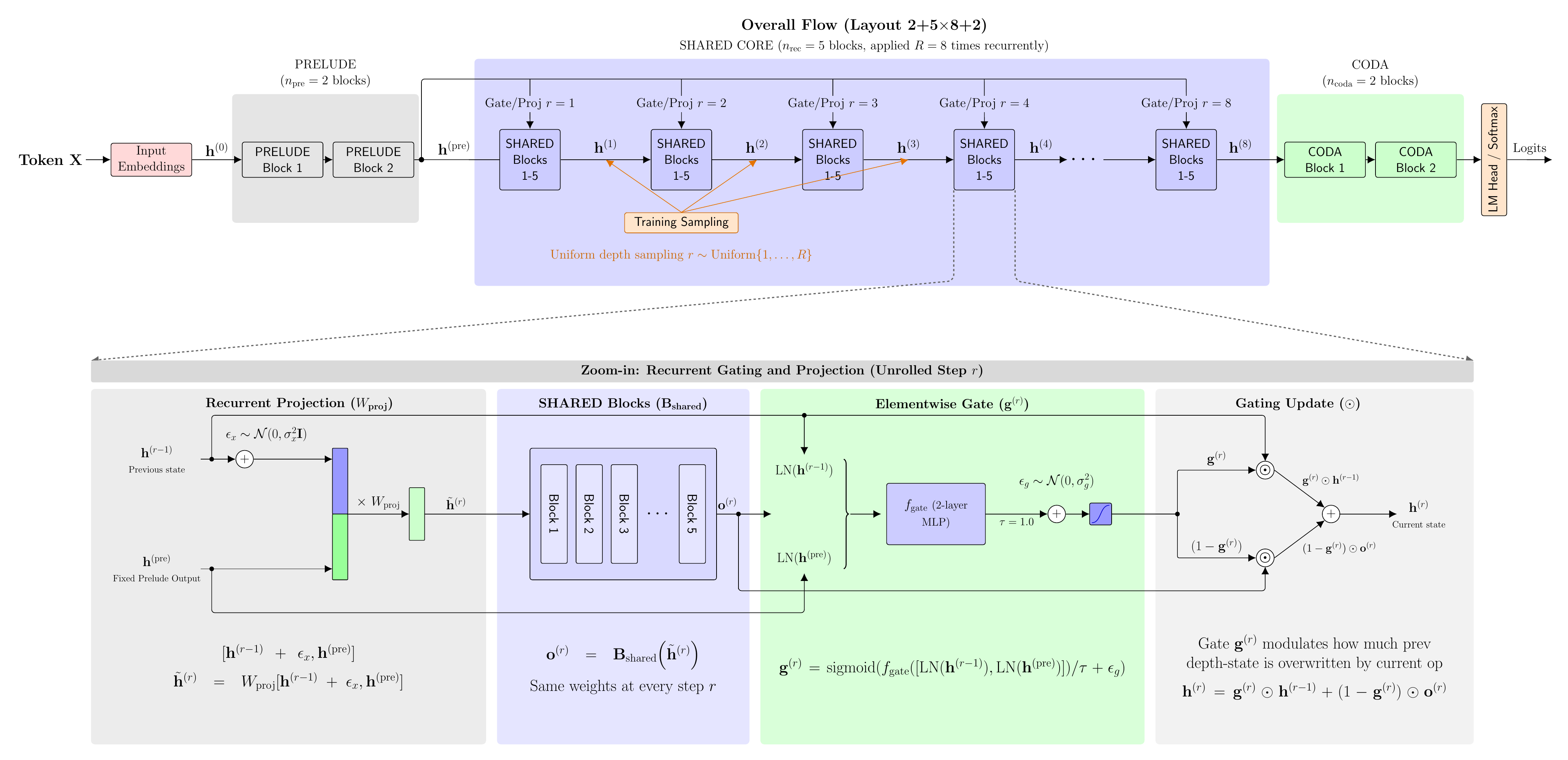}
  \caption{
    \textbf{\method{} architecture (\texttt{2+5$\times$8+2} layout).}
    \emph{Top:} Two prelude blocks (gray) encode a fixed context representation
    $\hstate^{(\mathrm{pre})}$ that is held constant across all recurrence steps.
    Five shared blocks (blue) are applied $R=8$ times recurrently; $\hstate^{(\mathrm{pre})}$
    conditions the projection and gate at each step. 
    Two coda blocks (green) project to logits.
    \emph{Bottom:} Zoom-in of a single recurrence step $r$: the current state
    $\hstate^{(r-1)}$ is projected together with $\hstate^{(\mathrm{pre})}$ via
    $W_{\mathrm{proj}}$, processed by the shared blocks to produce proposal
    $\mathbf{o}^{(r)}$, and blended back via the elementwise gate
    $\gate^{(r)} \in [0,1]^{T \times d}$.
  }
  \label{fig:architecture}
\end{figure}

\noindent
Total unique parameter count is $\Theta\bigl((\prelude + \nrec + \coda) \cdot d^2\bigr)$,
independent of $\recmax$.
A configuration such as \texttt{2+5$\times$4+2} visits $2 + 5 \cdot 4 + 2 = 24$
block executions per forward pass while storing weights for only $2 + 5 + 2 = 9$ blocks---
$2.6\times$ fewer unique blocks than the 24-layer GPT-2 medium baseline
it is isoFLOP-matched to.
\figref{fig:architecture} illustrates this layout.

Per-token forward-pass cost for a $\prelude \texttt{+} \nrec\!\times\!\recmax \texttt{+} \coda$
configuration at sequence length $S$ is
\begin{equation}
  \flops = \bigl(\prelude + \nrec \recmax + \coda\bigr)\bigl(24 d^2 + 4 S d\bigr)
           \;+\; \recmax \cdot 10 d^2 ,
  \label{eq:flops}
\end{equation}
where $24d^2 + 4Sd$ is the standard per-block cost (attention and MLP projections
plus the two attention matmuls) and $10 d^2$ is the per-step overhead of the
recurrent projection $W_{\mathrm{proj}}$ and the gate MLP $f_\gate$.
 
\subsection{Recurrent Projection and Elementwise Gate}
\label{subsec:gating}
 
Merely feeding the output of each recurrence as the input to the next recurrence risks
forgetting the original input context, suffers from vanishing or exploding gradients across
deep recurrences, and taxes the representational capacity of a single shared weight tensor.

Inspired by RNNs~\citep{ElmanRNN} in general, and GRUs~\citep{GRU} in particular, we treat the hidden representation
after each recurrence step as a \emph{state} that evolves across recurrences
(as illustrated in \figref{fig:teaser}).
Each application of the shared core reads this state, transforms it, and writes
an updated state back---analogous to an RNN cell, but operating over depth rather
than time.
This framing motivates the gating mechanism below: just as GRU gates control
how much of its cell state is overwritten at each timestep, \method{}'s elementwise gate
controls how much of the current depth-state is replaced by the shared block's output.

Hence, we introduce our novel adaptive gating approach that aims to mitigate the 3 aforementioned problems. At each recurrence step $r$, \method{} prepares the input to $\Bshared$
by projecting the current state together with the fixed prelude output,
then selectively writes the block's output back to the residual stream
via a learned elementwise gate.
The full update is:
 
\begin{align}
  \tilde{\hstate}^{(r)} &= W_{\mathrm{proj}}\,
    \bigl[\hstate^{(r-1)} + \epsilon_x,\;\hstate^{(\mathrm{pre})}\bigr],
    \label{eq:proj} \\
  \mathbf{o}^{(r)} &= \Bshared\!\left(\tilde{\hstate}^{(r)}\right),
    \label{eq:block} \\
  \hstate^{(r)} &= \gate^{(r)} \odot \hstate^{(r-1)}
                 + \bigl(1 - \gate^{(r)}\bigr) \odot \mathbf{o}^{(r)},
    \label{eq:gate_update}
\end{align}
where $[\cdot,\cdot]$ denotes concatenation along the feature axis,
$\gate^{(r)} \in [0,1]^{T \times d}$ is a learned elementwise gate (defined
in full below in Eq.~\eqref{eq:gate}),
$W_{\mathrm{proj}} \in \reals^{d \times 2d}$ is a learned projection,
$\epsilon_x \sim \mathcal{N}(0,\,\sigma^2 \mathbf{I})$ is injected additive
noise, and $\odot$ is
elementwise multiplication.

The gate $\gate^{(r)} \in [0,1]^{T \times d}$ is produced by a small
feed-forward network conditioned on the normalised current state and the
normalised prelude output:
\begin{equation}
  \gate^{(r)} = \sigmoid\!\left(
    f_\gate\!\bigl(
      [\mathrm{LN}(\hstate^{(r-1)}),\; \mathrm{LN}(\hstate^{(\mathrm{pre})})]\bigr)
    / \tau + \epsilon_g
  \right),
  \label{eq:gate}
\end{equation}
where $\sigmoid(\cdot) = 1/(1+e^{-(\cdot)})$ is the sigmoid function,
$\mathrm{LN}$ denotes layer normalization~\citep{ba2016layernormalization}, $\tau$ is a temperature hyperparameter
(set to $1.0$ throughout our experiments), $\epsilon_g \sim \mathcal{N}(0,\,\sigma_g^2)$
is per-scalar gate noise injected during training, and $f_\gate$ is a two-layer
MLP with SiLU activation and hidden dimension $d_\text{gate} = d$.
The second linear layer of $f_\gate$ has its bias initialised to $+4$, placing
$\gate^{(r)} \approx 0.98$ at the start of training: the copy branch of
\eqrefx{eq:gate_update} dominates, the residual stream passes through the
recurrence almost unchanged, and the model learns which elements to overwrite as
training proceeds. Positive initialisation of a forget gate is long-standing
practice in recurrent networks~\citep{gers2000learningtoforget}, and
\citet{jozefowicz2015empirical} found a bias of $+1$ or $+2$ enough to bring a
vanilla LSTM level with the best variants in their search. We simply apply the
same principle over depth rather than time. We swept across
$\{-2, 0, +2, +4\}$ at the full training horizon the spread is $0.019$ nats (\secref{app:sweeps}).
 
By defining $\gate^{(r)}_{t,i}$ as the $i^{\text{th}}$ element of the gate for token $t$ at recurrence $r$, \method{} achieves a highly granular, per-element specialization. When $\gate^{(r)}_{t,i} \to \mathbf{1}$, the residual stream remains unchanged, whereas $\gate^{(r)}_{t,i} \to \mathbf{0}$ allows the block output to fully replace the state. This per-recurrence specialization provides a more memory-efficient alternative to approaches such as Per-Layer Embeddings (PLE) (used in recent models like Gemma3n~\citep{google2025gemma3n} and Gemma-4~\citep{google2026gemma4}). While PLE requires $L$ unique embedding layers to provide layer-specific context, our approach conditions the gate on $\hstate^{(\mathrm{pre})}$ to provide a constant view of the \emph{original} input at every step, while modulating it with $\hstate^{r-1}$ and stochasticity $\epsilon_g$.

State noise $\epsilon_x$ (Eq.~\eqref{eq:proj}) perturbs the recurrent projection input,
discouraging the model from learning brittle exact-match patterns across steps.
Gate noise $\epsilon_g$ (Eq.~\eqref{eq:gate}) prevents the gate from collapsing to a
near-constant value during training.
Specific noise magnitudes ($\sigma_x$, $\sigma_g$) are reported in \secref{subsec:setup}.

\subsection{Recurrence Depth Sampling During Training}
\label{subsec:depth_sampling}
At each training step, we sample $r \sim \mathrm{Uniform}\{1, \ldots, R\}$. 
This serves two purposes: (1) it implicitly trains every exit point, enabling \textbf{early exit at inference without auxiliary loss terms} or the associated gradient interference common in multi-exit models~\citep{chen2023eellm, elhoushi-etal-2024-layerskip}; and (2) it acts as stochastic depth regularization, improving final validation loss over fixed-depth schedules (as will be shown later in our ablations~\secref{sec:ablations}.


We summarize the forward pass of \method{} in \listref{list:pseudocode}.

\section{Experiments \& Results}
\label{sec:experiments}

\paragraph{Experimental Setup}
\label{subsec:setup}

We build on the nanoGPT codebase~\citep{karpathy2022nanogpt} for our
transformer backbone implementation.
All models are trained on a diverse dataset of text with
sequence length $T = 1024$ tokens and GPT-2 BPE tokenisation (50{,}257-token vocabulary).
We use AdamW~\citep{loshchilov2019decoupled} with $\beta_1 = 0.9$,
$\beta_2 = 0.95$, weight decay $0.1$, a 2{,}000-step linear learning rate warmup, and a cosine learning-rate schedule
from a peak of $6\times10^{-4}$ to $6\times10^{-5}$.

Gradient norms are clipped at $1.0$. All runs used \texttt{bfloat16} mixed precision.
State noise $\sigma_x = 0.1$ and gate noise $\sigma_g = 0.1$ are used throughout all
\method{} runs; the gate temperature is fixed at $\tau = 1.0$.

We train for 20{,}000-steps and use an effective batch size of $\approx$491{,}520 tokens per step
(batch size 8, gradient accumulation 60, sequence length 1024), totalling
approximately \textbf{9.8B} tokens per run.
Dense baselines (GPT-2 small, medium, and large~\citep{gpt2}) are trained
from scratch under the identical setup; no pre-trained weights are used. We train four recurrent competitors from scratch under the same recipe: \emph{Mixture-of-Recursions} (MoR)~\citep{bae2026mixtureofrecursions}; a variant following \citet{huggin2024} that uses the prelude--coda layout with heavy-tail Poisson depth sampling; \emph{Relaxed Recursive Transformers} (RRT)~\citep{bae2025relaxed}, which relaxes weight tying with per-recurrence LoRA adapters; and \emph{Ouro}~\citep{ouro}, which supervises every loop iteration.


\begin{table}[t]
\centering
\caption{
  \textbf{Main results (validation loss).}
  \emph{Upper}: \isoflop{} regime---\method{} matches forward-pass FLOPs of the dense baseline with fewer unique parameters.
  \colorbox{gray!20}{Shaded rows} are dense baselines.
  \emph{Lower}: \isoparam{} regime---matched unique parameters, higher FLOPs per step.
  \textbf{Layers} = unique transformer blocks. $\downarrow$ lower is better.
  $^{\dagger}$RRT adds one LoRA adapter per recurrence, so its weights are not identical across steps.
}
\label{tab:main}
\small
\begin{tabular}{llrrrc}
\toprule
\textbf{Model} & \textbf{Config} & \textbf{Layers} & \textbf{FLOPs/fwd} & \textbf{Params} & \textbf{Val Loss $\downarrow$} \\
\midrule
\multicolumn{6}{l}{\emph{\isoflop{} regime (matched FLOPs per forward pass)}} \\
\midrule
\multicolumn{6}{l}{\textit{GPT-2 Small}} \\
\rowcolor{gray!15}
Baseline        & 12L                   & 12  & 1.84~G  & 124M  & 3.15 \\
MoR~\citep{bae2026mixtureofrecursions}            & 1+1$\times$10+1  &  3  & 1.84~G  & 45M   & 3.30 \\
Heavy-tail Poisson~\citep{huggin2024}             & 1+1$\times$10+1  &  3  & 1.84~G  & 34M   & 3.23 \\
Ouro~\citep{ouro}                                 & 1+1$\times$10+1  &  3  & 1.84~G  & 35M   & 3.19 \\
RRT~\citep{bae2025relaxed}$^{\dagger}$            & 1+1$\times$10+1  &  3  & 1.84~G  & 33M & 3.14 \\
\methodshort{} (ours)              & 1+1$\times$10+1       &  3  & 1.84~G  & 35M   & \textbf{3.14} \\
\addlinespace
\multicolumn{6}{l}{\textit{GPT-2 Medium}} \\
\rowcolor{gray!15}
Baseline       & 24L                   & 24  & 7.35~G  & 354M  & 2.84 \\
MoR~\citep{bae2026mixtureofrecursions}            & 2+5$\times$4+2  &  9  & 7.35~G  & 137M  & 3.02 \\
Heavy-tail Poisson~\citep{huggin2024}             & 2+5$\times$4+2  &  9  & 7.35~G  & 124M  & 2.97 \\
Ouro~\citep{ouro}                                 & 2+5$\times$4+2  &  9  & 7.35~G  & 126M  & 2.93 \\
RRT~\citep{bae2025relaxed}$^{\dagger}$            & 2+5$\times$4+2  &  9  & 7.35~G  & 125M  & 2.95 \\
\methodshort{} (ours)              & 2+5$\times$4+2        &  9  & 7.35~G  & 127M  & \textbf{2.89} \\
\addlinespace
\multicolumn{6}{l}{\textit{GPT-2 Large}} \\
\rowcolor{gray!15}
Baseline        & 36L                   & 36  & 21.1~G  & 774M  & 2.71 \\
MoR~\citep{bae2026mixtureofrecursions}            & 1+5$\times$6+5  & 11  & 21.1~G  & 309M  & 2.91 \\
Heavy-tail Poisson~\citep{huggin2024}             & 1+5$\times$6+5  & 11  & 21.1~G  & 291M  & 2.93 \\
Ouro~\citep{ouro}                                 & 1+5$\times$6+5  & 11  & 21.1~G  & 290M  & 2.82 \\
RRT~\citep{bae2025relaxed}$^{\dagger}$            & 1+5$\times$6+5  & 11  & 21.1~G  & 289M  & 2.85 \\
\methodshort{} (ours)              & 1+5$\times$6+5        & 11  & 21.1~G  & 293M  & \textbf{2.77} \\
\midrule
\multicolumn{6}{l}{\emph{\isoparam{} regime (matched unique parameter count)}} \\
\midrule
\multicolumn{6}{l}{\textit{GPT-2 Small}} \\
\rowcolor{gray!15}
Baseline        & 12L                   & 12  & 1.84~G  & 124M  & 3.15 \\
MoR~\citep{bae2026mixtureofrecursions}            & 1+10$\times$10+1  &  12  & 15.64~G  & 135M   & 3.12 \\
Heavy-tail Poisson~\citep{huggin2024}             & 1+10$\times$10+1  &  12  & 15.64~G  & 124M   & 3.06 \\
Ouro~\citep{ouro}                                 & 1+10$\times$10+1  &  12  & 15.64~G  & 127M   & 3.10 \\
RRT~\citep{bae2025relaxed}$^{\dagger}$            & 1+10$\times$10+1  &  12  & 15.64~G  & 125M & 3.08 \\
\methodshort{} (ours)              & 1+10$\times$10+1       &  12  & 15.64~G  & 127M   & \textbf{3.04} \\
\addlinespace
\multicolumn{6}{l}{\textit{GPT-2 Medium}} \\
\rowcolor{gray!15}
Baseline  & 24L                   & 24  & 7.35~G  & 354M  & 2.84 \\
MoR~\citep{bae2026mixtureofrecursions}              & 2+20$\times$4+2       & 24  & 26.4~G  & 367M  & 2.78 \\
Heavy-tail Poisson~\citep{huggin2024}             & 2+20$\times$4+2       & 24  & 26.4~G  & 356M  & \textbf{2.74} \\
\methodshort{} (ours)              & 2+20$\times$4+2       & 24  & 26.4~G  & 357M  & 2.76 \\
\addlinespace
\multicolumn{6}{l}{\textit{GPT-2 Large}} \\
\rowcolor{gray!15}
Baseline  & 36L                   & 36  & 21.1~G  & 774M  & 2.71 \\
MoR~\citep{bae2026mixtureofrecursions}            & 3+30$\times$6+3       & 36  & 109.0~G & 795M  & 2.69 \\
Heavy-tail Poisson~\citep{huggin2024}             & 3+30$\times$6+3       & 36  & 109.0~G & 776M  & 2.70 \\
\methodshort{} (ours)              & 3+30$\times$6+3       & 36  & 109.0~G & 779M  & \textbf{2.65} \\
\bottomrule
\end{tabular}
\end{table}


\definecolor{isoflop_col}{RGB}{230,100,30}   
\definecolor{isoparam_col}{RGB}{30,90,200}    

\newcommand{\isoflophigh}[1]{\textcolor{isoflop_col}{\textbf{#1}}}
\newcommand{\isoparamhigh}[1]{\textcolor{isoparam_col}{\textbf{#1}}}
\newcommand{\dpos}[1]{\textcolor{black!60}{$\uparrow$#1}}
\newcommand{\dneg}[1]{\textcolor{black!60}{$\downarrow$#1}}

\begin{table}[t]
\centering
\caption{
  \textbf{Downstream task evaluation (large scale).}
  Zero-shot accuracy on standard benchmarks using \texttt{lm-eval-harness}.
  $\uparrow$ higher is better; deltas relative to GPT-2 Large.
  \textcolor{isoflop_col}{\textbf{Orange}}: best \isoflop{} result;
  \textcolor{isoparam_col}{\textbf{blue}}: best \isoparam{} result.
  The \isoparam{} model outperforms the dense baseline on average by \textbf{+2.10 points};
  the \isoflop{} model matches GPT-2 Large at only 37\% of its parameter count.
}
\label{tab:downstream_large}
\small
\begin{tabular}{lccc}
\toprule
\multirow{3}{*}{\textbf{Task}} &
\textbf{\methodshort{} \isoflop{}} &
\textbf{GPT-2 Large} &
\textbf{\methodshort{} \isoparam{}} \\
& \textbf{1+5$\times$6+5} & \textbf{(baseline)} & \textbf{3+30$\times$6+3} \\
& 288M, 11L ($-$63\%) & 774M, 36L & 774M, 36L \\
\midrule
ARC-Challenge ($\uparrow$)      & \isoflophigh{25.43} \dpos{1.80} & 23.63 & 25.26 \dpos{1.63}               \\
ARC-Easy ($\uparrow$)           & 42.89 \dneg{0.42}               & 43.31 & \isoparamhigh{45.54} \dpos{2.23} \\
BoolQ ($\uparrow$)              & \isoflophigh{60.18} \dpos{2.60} & 57.58 & 52.29 \dneg{5.29}               \\
HellaSwag ($\uparrow$)          & 35.62 \dneg{1.59}               & 37.21 & \isoparamhigh{41.52} \dpos{4.31} \\
LAMBADA (OpenAI) ($\uparrow$)   & 39.05 \dneg{1.00}               & 40.05 & \isoparamhigh{45.27} \dpos{5.22} \\
LAMBADA (Standard) ($\uparrow$) & 30.60 \dpos{0.87}               & 29.73 & \isoparamhigh{38.02} \dpos{8.29} \\
OpenBookQA ($\uparrow$)         & 28.40 \dneg{1.40}               & 29.80 & \isoparamhigh{30.20} \dpos{0.40} \\
PIQA ($\uparrow$)               & 64.31 \dneg{1.20}               & 65.51 & \isoparamhigh{66.05} \dpos{0.54} \\
Winogrande ($\uparrow$)         & 52.25 \dpos{0.63}               & 51.62 & \isoparamhigh{53.28} \dpos{1.66} \\
\midrule
Average ($\uparrow$)            & \isoflophigh{42.08} \dpos{0.03} & 42.05 & \isoparamhigh{44.15} \dpos{2.10} \\
\bottomrule
\end{tabular}
\end{table}

\paragraph{Main Results}
Table~\ref{tab:main} summarizes performance across model sizes. In the \textbf{\isoflop{} regime}, \method{} achieves competitive quality while storing only \textbf{36--37\%} of the baseline's parameters. At small scale it reaches lower loss than the dense baseline on all three seeds we ran: $3.145 \pm 0.004$ against $3.188 \pm 0.056$, our worst seed at $3.148$ and the baseline's best at $3.154$ (Appendix~\ref{app:seeds}). At medium and large scale the dense baseline leads at the standard budget, by $0.05$ and $0.06$ nats respectively, and \secref{subsec:abl_data} shows both gaps closing as the token budget grows.

\begin{figure}[t]
  \centering
  \begin{subfigure}[t]{0.48\linewidth}
    \includegraphics[width=\linewidth]{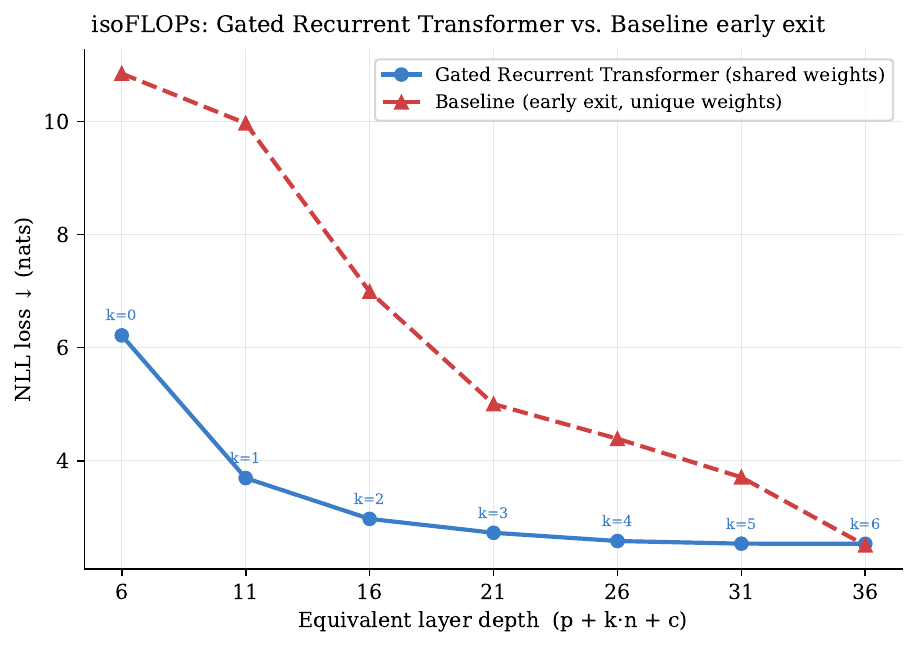}
    \caption{%
      Early-exit Validation loss vs.\ equivalent layer depth $p + k \cdot n + c$.
    }
    \label{fig:isoflops}
  \end{subfigure}
  \hfill
  \begin{subfigure}[t]{0.48\linewidth}
    \includegraphics[width=\linewidth]{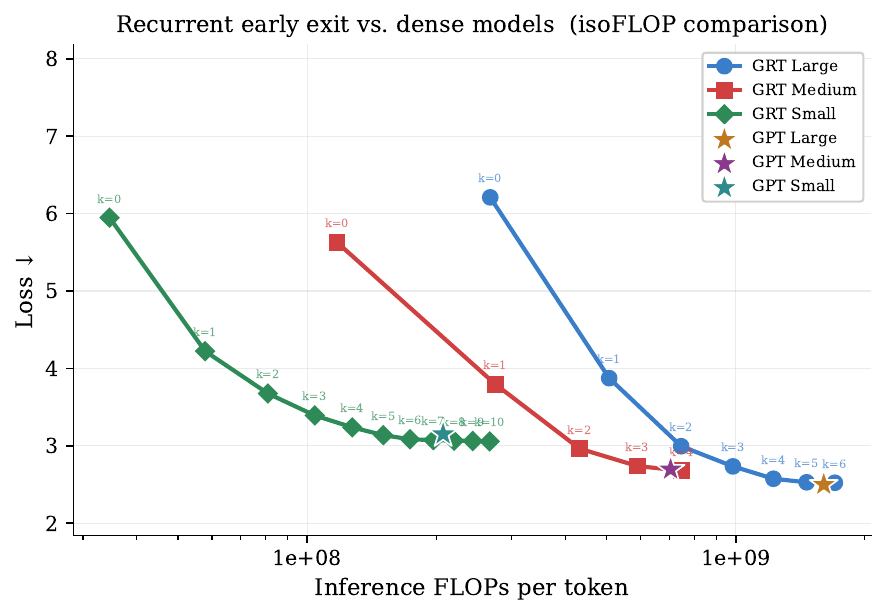}
    \caption{%
      Early-exit Validation loss vs.\ inference FLOPs per token across Small, Medium,
      and Large scales at varying recurrence depths $k$.
    }
    \label{fig:flops_loss}
  \end{subfigure}
  \caption{%
    \textbf{Early-Exit Analysis} Inference FLOPs are measured based on sequence length 1024,
      batch size 1.
  }
  \label{fig:compute_tradeoff}
\end{figure}

Against the recurrent baselines the margin widens with scale. RRT is the strongest of the four and reaches parity at small scale, $3.143$ against our $3.141$, a difference well inside the $\pm 0.004$ seed spread; at medium it trails by $0.06$ nats and at large by $0.08$. MoR and heavy-tail Poisson trail by $0.08$ to $0.16$ nats at every scale. We attribute the widening gap to what each method holds fixed. RRT's LoRA deltas are chosen at training time and applied identically to every input, so the diversity they buy is fixed in advance and does not grow with the number of recurrences, whereas the gate conditions on the state it is about to update and keeps successive states separable where uniform sharing collapses them. Ouro is competitive at small scale ($3.19$) but its medium run converged to $2.93$ and its large run converged to $2.82$. Strict weight tying is also what leaves $\recinf$ free at inference and produces the early-exit curve of \figref{fig:isoflops}, which a fixed set of adapters cannot produce from a single checkpoint. Appendix~\ref{app:method_comparison} tabulates the design differences alongside training cost.

In the \textbf{\isoparam{} regime}, increasing recurrence $R$ at a fixed parameter budget yields significant gains: a \textbf{0.08} nat improvement at Medium scale and \textbf{0.06} at Large scale. This demonstrates that for a fixed memory footprint, trading inference FLOPs for recurrent depth is a principled strategy for improving model expressivity.

Table~\ref{tab:downstream_large} reports accuracy across nine benchmarks at
large scale using \texttt{lm-eval-harness}~\citep{eval-harness}.
The \isoflop{} variant matches the dense model on average (42.08 vs.\ 42.05),
confirming that parameter efficiency translates across evaluation protocols, while
the \isoparam{} variant outperforms the dense baseline
on eight of nine tasks, consistent with the validation-loss advantage in \tabref{tab:main}. 

\paragraph{Early Exit Analysis}
As shown in Fig.~\ref{fig:isoflops}, \method{} exhibits superior early-exit performance as an emergent property: at matched inference FLOPs \method{} exiting with fewer recurrences has better loss than a dense model exiting at an earlier layer. Fig.~\ref{fig:flops_loss} extends the analysis across model scales. Because uniform depth sampling (\secref{subsec:depth_sampling}) trains all intermediate recurrent states to predict final losses, the model provides a continuous ``compute--quality dial''. This allows a single checkpoint to transition from fast, shallow inference to high-quality deep inference without re-training or auxiliary losses. 

\begin{figure}[h]
  \centering
  \includegraphics[width=\linewidth]{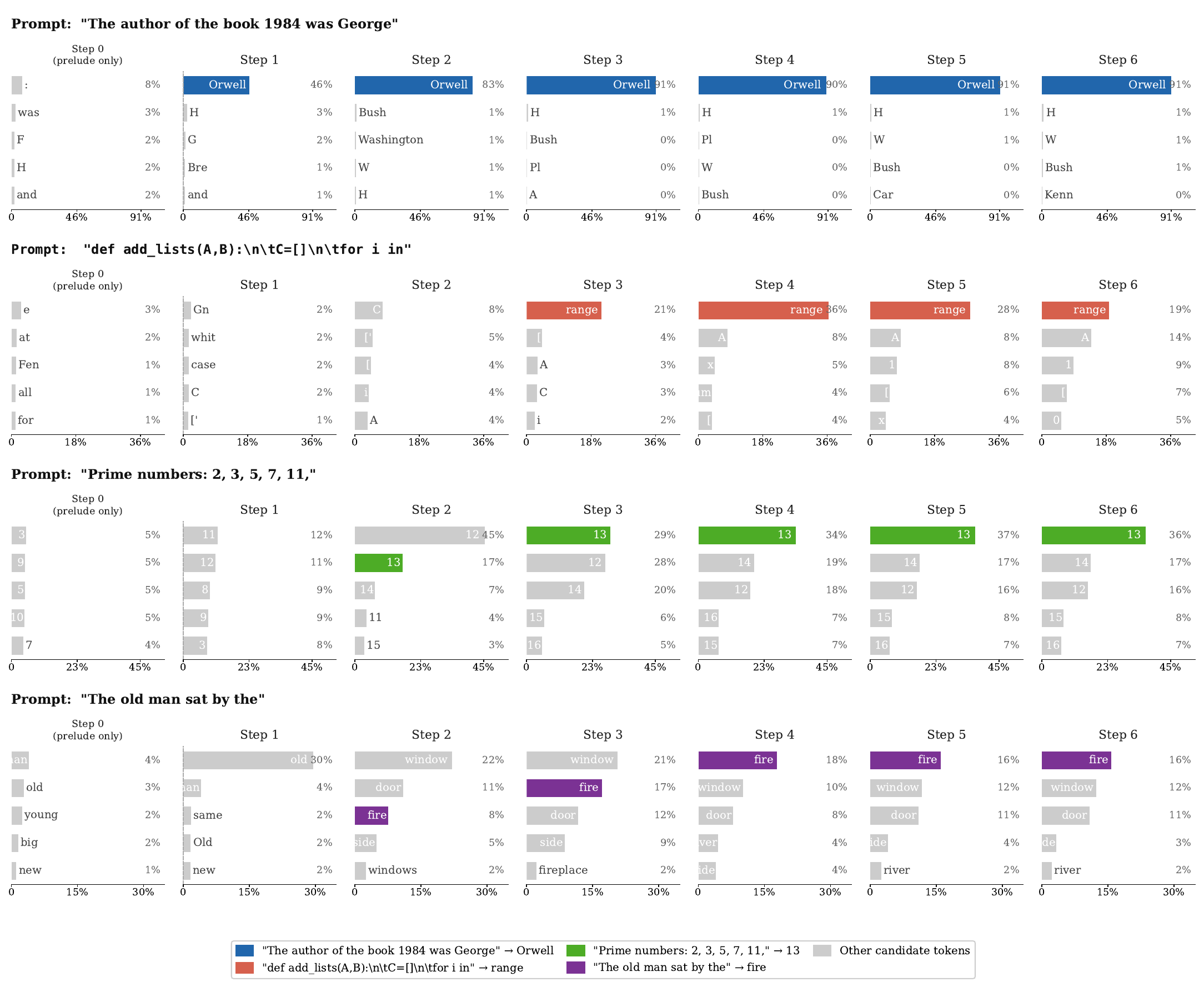}
  \caption{
    \textbf{Next-token probability at each recurrence step.}
    Each row shows one prompt; the correct answer is highlighted in colour.
    Prelude-only output (leftmost column) captures partial signal;
    by step~2 the correct token dominates in three of four prompts.
    Step labels use 1-indexed notation consistent with the analysis section.
  }
  \label{fig:probing_barcharts}
\end{figure}

In \figref{fig:probing_barcharts}, we provide a qualitative analysis of how next-token predictions evolve across recurrence steps. Our findings reveal that \method{} adapts its computational depth to the semantic demands of the prompt: (1) \textbf{Knowledge-intensive} tokens often stabilize early in the recurrence trajectory; (2) \textbf{Reasoning-based} tasks demonstrate progressive error correction and sharpening of the probability mass over deeper iterations; and (3) \textbf{Open-ended or creative} prompts exhibit continuous linguistic refinement, where deeper recurrence enhances local coherence despite the lack of a singular ground truth. This suggests that the model implicitly utilizes the recurrent core to modulate its internal ``thinking time'' based on task complexity.

\paragraph{KV Cache Sharing}
\label{subsec:kvcache}

Naively, \method{} requires $\recmax$ separate KV caches during decoding---one per
recurrence step---multiplying memory by $\recmax$ over a standard transformer.
Following~\citep{ouro}, we evaluate three compressed strategies that each reduce recurrent KV memory
over $\recmax$ layers, equivalent to $3.27\times$ reduction across the model:
reusing only the \emph{last} step's cache, only the \emph{first} step's, or
an \emph{averaged} K/V across steps.
As shown in \figref{fig:kvcache_bar}, we observe that \emph{averaged} leads to best accuracy, while \emph{last} leads to the worst accuracy.

\paragraph{Wall-Clock Latency and Decoding Memory}
\label{subsec:latency}

\method{} executes the
same number of block applications as the dense model it is \isoflop{}-matched to,
distributed over fewer unique weight tensors.
\tabref{tab:latency} reports measured generation latency and peak GPU memory on
identical hardware (batch size 4, prompt length 1024, 128 generated tokens).
Under \texttt{torch.compile}, \method{}-Large costs $+10\%$ generation latency for
$62\%$ fewer parameters and $59\%$ less peak memory; at medium scale the overhead
is $+11\%$. In eager mode both overheads are $+23\%$, so roughly half the gap is
kernel-launch overhead from the additional elementwise operations rather than
arithmetic.

\begin{table}[h]
  \centering
  \caption{%
    \textbf{Generation latency and peak memory (\isoflop{} configurations).}
    Batch size 4, prompt length 1024, 128 generated tokens, same hardware.
    Both arms are measured under \texttt{torch.compile} and in eager mode.
  }
  \label{tab:latency}
  \small
  \setlength{\tabcolsep}{6pt}
  \begin{tabular}{llrrrr}
    \toprule
    \textbf{Model} & \textbf{Config} & \textbf{Params} &
    \textbf{Eager (ms/tok)} & \textbf{Compiled (ms/tok)} & \textbf{Peak mem.} \\
    \midrule
    GPT-2 Medium      & 24L                  & 354M & 2.95 & 2.54 & 747~MB \\
    \methodshort{} Medium  & \texttt{2+5$\times$4+2} & 127M & 3.62 & 2.81 & \textbf{402~MB} \\
    \addlinespace
    GPT-2 Large       & 36L                  & 774M & 4.33 & 3.33 & 1570~MB \\
    \methodshort{} Large   & \texttt{1+5$\times$6+5} & 293M & 5.33 & 3.67 & \textbf{639~MB} \\
    \bottomrule
  \end{tabular}
\end{table}

The $R\times$-expanded KV cache is a real cost, and whether it erodes the parameter
saving depends on batch size.
\tabref{tab:decode_mem} reports end-to-end decoding memory (weights $+$ KV, bf16,
$T=1024$) for the medium \isoflop{} checkpoint.
At $B{=}1$ decoding is weight-dominated and the parameter reduction lands at
$0.55\times$ dense memory even with the naive $R\times$ cache; at $B{=}32$ the cache
dominates and the naive strategy recovers only $0.91\times$.
Averaging K/V across recurrence steps brings this to $0.39\times$ while
\emph{improving} HellaSwag accuracy over the full cache ($33.90$ vs.\ $33.65$),
which suggests the averaging acts as a mild regulariser at this scale.

\begin{table}[h]
  \centering
  \caption{%
    \textbf{End-to-end decoding memory (medium \isoflop{} checkpoint, bf16, $T=1024$).}
    Weights $+$ KV cache. Parenthesised values are relative to GPT-2 Medium.
  }
  \label{tab:decode_mem}
  \small
  \setlength{\tabcolsep}{8pt}
  \begin{tabular}{lccc}
    \toprule
    \textbf{Configuration} & \textbf{HellaSwag} & \textbf{Mem.\ @ $B{=}1$} & \textbf{Mem.\ @ $B{=}32$} \\
    \midrule
    GPT-2 Medium (24L, 354M)          & \textbf{34.45} & 0.75~GiB (1.00$\times$) & 3.66~GiB (1.00$\times$) \\
    \methodshort{} \isoflop{}, full $R\times$ KV & 33.65 & 0.41~GiB (0.55$\times$) & 3.32~GiB (0.91$\times$) \\
    \methodshort{} \isoflop{}, averaged KV      & 33.90 & \textbf{0.35~GiB (0.47$\times$)} & \textbf{1.44~GiB (0.39$\times$)} \\
    \bottomrule
  \end{tabular}
\end{table}

\paragraph{Mechanistic Analysis Summary}
\label{subsec:analysis_summary}

Detailed mechanistic analysis of the large \texttt{1+5$\times$6+5} checkpoint
is provided in Appendix~\ref{app:analysis}; we summarise the key findings here.
We observe in \figref{fig:gate_behaviour} that gate, $g^{(r)}$, behavior evolves across recurrence steps: early recurrences
($r \leq 2$) are \emph{write-heavy} (gate values near $0$, block output dominates),
while later steps transition to \emph{copy-heavy} behavior (gate near $1$, residual
stream preserved), consistent with a coarse-to-fine computation strategy.
Centered Kernel Alignment~\citep{kornblith2019similarity} analysis in \figref{fig:cka} confirms that
representations across recurrence steps are more similar to one another than
representations across layers in a standard transformer of matched depth, suggesting
the shared block learns a broadly applicable transformation rather than
depth-specialized ones.
In \figref{fig:token_difficulty}, we observe that tokens that are more difficult for the model accumulate $10\times$ more improvement across recurrence steps, demonstrating implicit compute routing toward harder inputs without any explicit difficulty signal.

\section{Ablations}
\label{sec:ablations}

We study two complementary aspects of \method{}: the contribution of each
architectural component (\secref{subsec:abl_components}) and the model's
behaviour as the training data budget varies (\secref{subsec:abl_data}).
All ablations use validation loss as the primary metric.

\paragraph{Component Ablation}
\label{subsec:abl_components}

\begin{figure}[t]
  \centering
  \includegraphics[width=0.56\linewidth]{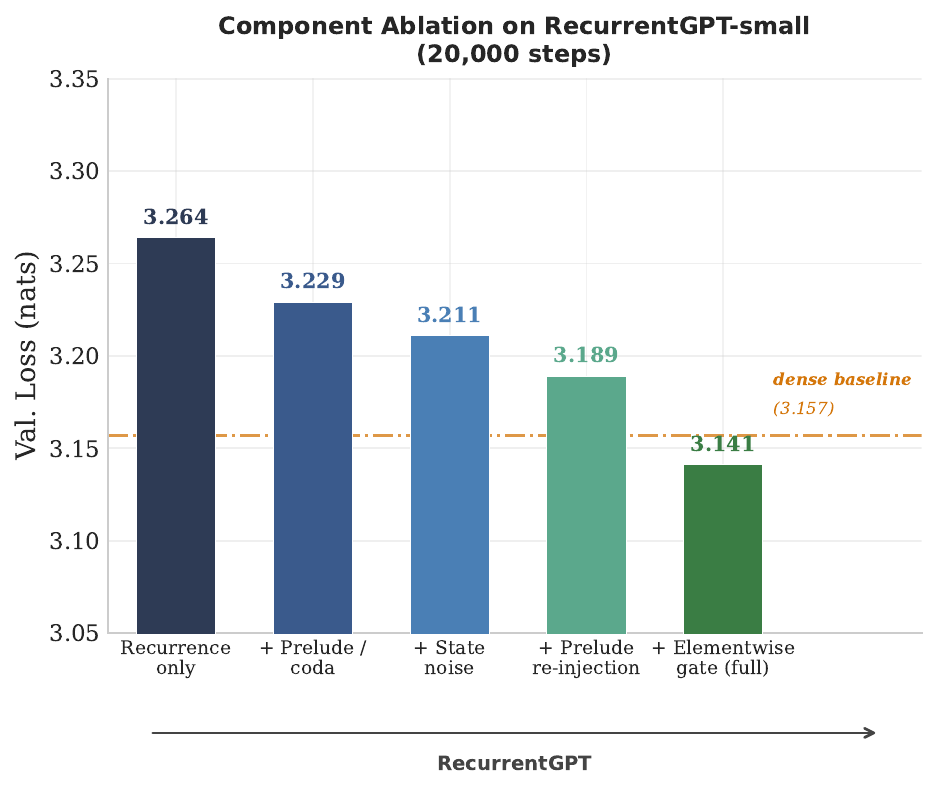}
  \caption{%
    \textbf{Component ablation on \method{}-small (20{,}000 steps).}
    Each bar adds one component to the one on its left. The dashed line is the dense
    12-layer baseline at $3.157$, trained identically; only the full model falls below
    it. The 2{,}000-step counterpart on the medium configuration is in
    Appendix~\ref{app:ablation_2k}.
  }
  \label{fig:ablation_ladder}
\end{figure}

\figref{fig:ablation_ladder} traces a sequential ablation on the small model at the full 20{,}000-step budget.
Recurrence alone raises validation loss by 0.107 nats: without a mechanism to differentiate steps,
the shared block conflates early and late processing into a single weight tensor.
Adding prelude and coda boundaries (row~3) recovers 0.035 nats; state noise (row~4) contributes 0.018 more,
which also isolates the value of injecting noise at every step rather than only at initialisation.
Prelude re-injection into each recurrence (row~5) cuts loss by a further 0.022 nats.
The elementwise gate (row~6) is the single largest contributor at $-$\textbf{0.048 nats}.



\begin{figure*}[t]
  \centering
  \begin{minipage}[t]{0.64\linewidth}
    \centering
    \includegraphics[width=\linewidth]{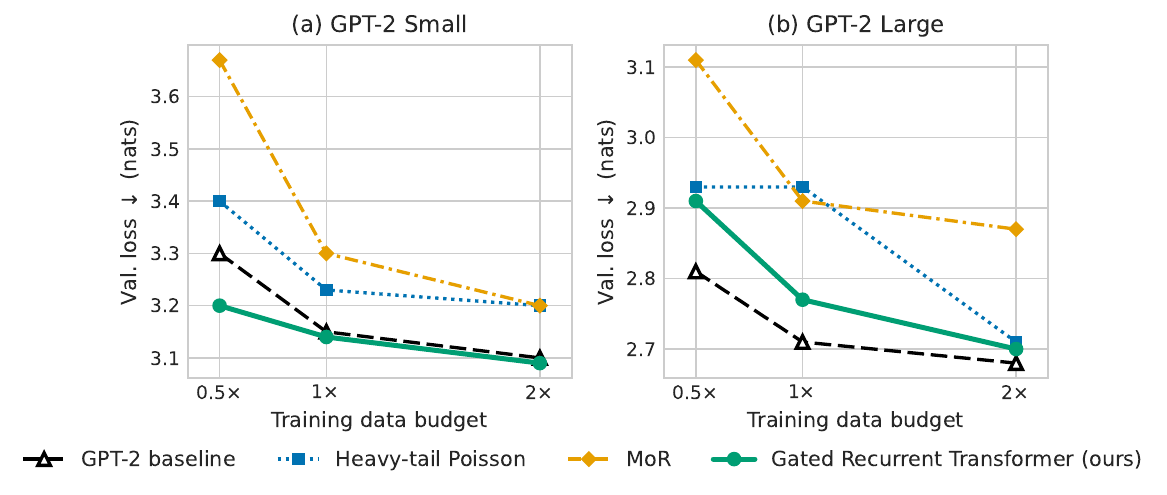}
    \caption{%
      \textbf{Validation loss vs.\ training data budget.}
      Validation loss vs.\ training data budget at small and large scale.
    }
    \label{fig:data_scaling}
  \end{minipage}
  \hfill
  \begin{minipage}[t]{0.33\linewidth}
    \centering
    \includegraphics[width=\linewidth]{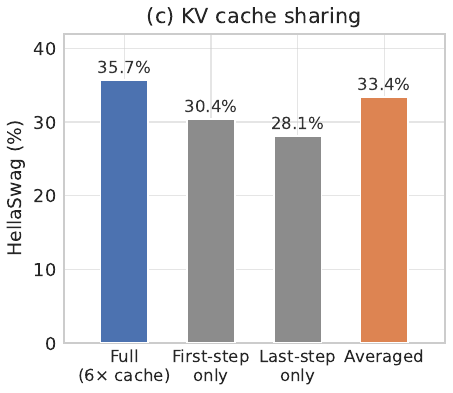}
    \caption{%
      \textbf{KV cache sharing.} HellaSwag accuracy (\%) for the
    large \isoflop{} checkpoint.
    }
    \label{fig:kvcache_bar}
  \end{minipage}
\end{figure*}

\paragraph{Data Efficiency}
\label{subsec:abl_data}

As illustrated in \figref{fig:data_scaling}, we analyze the performance of recurrent depth across varying training data budgets. At smaller model scales, \method{} consistently maintains a superior Pareto frontier compared to the dense baseline across all observed data volumes. At larger scales, a more complex ``crossover'' dynamic emerges: while the dense model shows an initial lead under low-data regimes, the gap narrows significantly as the token budget increases. For the Large-scale configuration, the trajectories converge at higher data budgets, suggesting that recurrent sharing is particularly effective at saturating parameter capacity when provided with sufficient data.



\section{Conclusion}
\label{sec:conclusion}
\method{} demonstrates that a learned per-element gate conditioning on the current
hidden state, the fixed prelude output, and injected stochastic noise is a
principled strategy for decoupling model depth from parameter count.
Under an \isoflop{} constraint, \method{} stores only $36$--$37\%$ of the dense
baseline's unique parameters while beating both MoR~\citep{bae2026mixtureofrecursions}
and heavy-tail Poisson sampling at every scale; under an \isoparam{} constraint,
deeper recurrence yields consistent validation-loss gains (Table~1) and downstream
improvements of up to $+8.29$ points (Table~2).
Three limitations bound the current work: the recurrence depth $R$ is fixed at
inference with no per-token halting~\citep{graves2017adaptivecomputationtimerecurrent};
the gate bias and noise magnitudes may require re-tuning
beyond the GPT-2 family; and the scale-dependent optimal sharing fraction warrants
systematic study.
Future directions include dynamic per-token halting and knowledge distillation
from a dense language model into a recurrent student.

\newpage
\bibliographystyle{unsrtnat}
\bibliography{main}

\appendix
\newpage

\section{Implementation Details}
\label{app:impl}

\begin{listing}
\begin{Verbatim}[fontsize=\small, frame=single, framesep=4pt,
                 xleftmargin=0pt, xrightmargin=0pt,
                 commandchars=\\\{\}]
\kw{def} grt_forward(x, model, R, r_min=1):      \cmt{# R = max recurrence depth}
    h = token_embed(x) + pos_embed(x)             \cmt{# [S, d]}
    \kw{for} block \kw{in} model.prelude_blocks:
        h = h + block(h)
    h_pre = h                                   \cmt{# h^(0) = h^(pre): frozen anchor}
    r = R \kw{if} \kw{not} training \kw{else} uniform_sample(r_min, R)  \cmt{# depth sampling}
    \kw{for} step \kw{in} range(r):                \cmt{# r in {0,...,R-1}}
        eps = sample_gaussian(0, sigma**2)        
        h_tilde = W_proj( cat([h + eps, h_pre]) ) \cmt{# concat -> project to d}
        o = B_shared(h_tilde)                     \cmt{# same weights every step}
        eps_g = sample_gaussian(0, sigma_g**2)
        g = sigmoid(f_gate(LN(h), LN(h_pre))/tau + eps_g) \cmt{# gate in [0,1]\^{}\{Sxd\}}
        h = g * h + (1 - g) * o                  \cmt{# gated residual update}
    \kw{for} block \kw{in} model.coda_blocks:
        h = h + block(h)
    \kw{return} lm_head(h)                        \cmt{# [S, |V|]}
\end{Verbatim}
\caption{
  \textbf{\method{} forward pass.}
}
\label{list:pseudocode}
\end{listing}

\subsection{Hyperparameter Table}

Table~\ref{tab:hyperparams} lists the full training configuration for every run
reported in the main paper.

\begin{table}[h]
\centering
\caption{%
  \textbf{Training configurations.}
  $p$: prelude blocks; $b$: shared blocks; $R$: recurrence steps; $c$: coda blocks;
  $d$: embedding dimension; $h$: attention heads.
  Batch tokens = batch\_size $\times$ gradient\_accumulation $\times$ sequence\_length $\times$ num\_GPUs.
  All runs use AdamW ($\beta_1=0.9$, $\beta_2=0.95$), weight decay 0.1, gradient clip 1.0,
  bfloat16 mixed precision, cosine LR schedule.
}
\label{tab:hyperparams}
\small
\setlength{\tabcolsep}{3pt}
\begin{tabular}{lcccccccccc}
\toprule
\textbf{Run} & $p$ & $b$ & $R$ & $c$ & $d$ & $h$ & \textbf{Batch tokens} & \textbf{Peak LR} & \textbf{Steps} & \textbf{Hardware} \\
\midrule
GPT-2 Small (baseline)       & --- & --- & --- & --- & 768  & 12 & 491,520 & $6\times10^{-4}$ & 20k & 2$\times$ H200 \\
\methodshort{} Small (isoFLOP)    & 1   & 1   & 10  & 1   & 768  & 12 & 491,520 & $6\times10^{-4}$ & 20k & 2$\times$ H200 \\
\addlinespace
GPT-2 Medium (baseline)      & --- & --- & --- & --- & 1024 & 16 & 491,520 & $6\times10^{-4}$ & 20k & 2$\times$ H200 \\
\methodshort{} Medium (isoFLOP)   & 2   & 5   & 4   & 2   & 1024 & 16 & 491,520 & $6\times10^{-4}$ & 20k & 2$\times$ H200 \\
\methodshort{} Medium (isoParam)  & 2   & 20   & 4  & 2   & 1024 & 16 & 491,520 & $6\times10^{-4}$ & 20k & 2$\times$ H200 \\
\addlinespace
GPT-2 Large (baseline)       & --- & --- & --- & --- & 1280 & 20 & 491,520 & $6\times10^{-4}$ & 20k & 2$\times$ H200 \\
\methodshort{} Large (isoFLOP)    & 1   & 5   & 6   & 5   & 1280 & 20 & 491,520 & $6\times10^{-4}$ & 20k & 2$\times$ H200 \\
\methodshort{} Large (isoParam)   & 3   & 30   & 6  & 3   & 1280 & 20 & 491,520 & $6\times10^{-4}$ & 20k & 2$\times$ H200 \\
\bottomrule
\end{tabular}
\end{table}

All recurrent runs use gate bias initialisation $+4$ ($\gate \approx 0.98$ at init),
state noise $\sigma_x = 0.1$, gate noise $\sigma_g = 0.1$, and gate temperature $\tau = 1.0$.
    During training, recurrence depth is sampled uniformly from $\{1,\ldots,R\}$ at each training step.

\section{Additional Ablations}
\label{app:ablations}

\subsection{Run-to-Run Variance Across Seeds}
\label{app:seeds}

\tabref{tab:seeds} reports three independent training runs (seeds 1337, 42, 123)
of the small \isoflop{} configuration and of its dense counterpart, all at the
full 20{,}000-step budget.
\method{} averages $3.145 \pm 0.004$ against $3.188 \pm 0.056$ for the dense
baseline.
\method{}'s worst seed
($3.148$) still falls below the dense baseline's best ($3.154$).
Seed 1337 is the run reported throughout the main paper.

\begin{table}[h]
  \centering
  \caption{%
    \textbf{Three-seed validation loss on the small configuration}
    (20{,}000 steps).
    Both models use the \isoflop{} setting of \tabref{tab:main}.
    Lower is better.
  }
  \label{tab:seeds}
  \small
  \setlength{\tabcolsep}{7pt}
  \begin{tabular}{lcccc}
    \toprule
    \textbf{Model} & \textbf{Seed 1337} & \textbf{Seed 42} & \textbf{Seed 123}
      & \textbf{Mean $\pm$ std} \\
    \midrule
    Dense baseline (12L)         & 3.154 & 3.156 & 3.253 & $3.188 \pm 0.056$ \\
    \methodshort{} (\texttt{1+1$\times$10+1}) & \textbf{3.141} & \textbf{3.146}
      & \textbf{3.148} & $\mathbf{3.145 \pm 0.004}$ \\
    \bottomrule
  \end{tabular}
\end{table}

\subsection{Component Ablation at Short Horizon}
\label{app:ablation_2k}

\tabref{tab:ablation_ladder_2k} reports the sequential ablation of
\secref{subsec:abl_components} run on the \emph{medium} configuration at
2{,}000 steps.
At this horizon the ordering of the last two rows is inverted relative to the
full 20{,}000-step run: prelude re-injection contributes $-0.198$ nats against
the gate's $-0.115$. We report both because the comparison is itself
informative --- the structural components land early, whereas the gate is a
learned mechanism whose contribution accrues over training.

\begin{table}[h]
  \centering
  \caption{%
    \textbf{Component ablation on \method{}-medium at 2\,000 steps.}
    Short-horizon counterpart to \figref{fig:ablation_ladder}.
    The dense baseline (row~1) is a standard 24-layer GPT trained identically.
  }
  \label{tab:ablation_ladder_2k}
  \small
  \setlength{\tabcolsep}{6pt}
  \begin{tabular}{clcc}
    \toprule
    & \textbf{Configuration} & \textbf{Val.\ loss} & $\boldsymbol{\Delta}$ \\
    \midrule
    1 & Dense baseline (non-recurrent) & 3.621 & --- \\
    \midrule
    2 & Recurrence         & 4.163 & $+$0.542 \\
    3 & \quad + Prelude / coda ($2{+}\_+2$)   & 4.118 & $-$0.045 \\
    4 & \quad + State noise ($\sigma_x{=}0.1$)  & 4.099 & $-$0.019 \\
    5 & \quad + Prelude re-injection          & 3.901 & $-$\textbf{0.198} \\
    6 & \quad + Elementwise gate (full \method) & \textbf{3.786} & $-$0.115 \\
    \bottomrule
  \end{tabular}
\end{table}

\subsection{Design Comparison of Recurrent-Depth Methods}
\label{app:method_comparison}

\tabref{tab:method_comparison} sets the small-scale results of \tabref{tab:main} beside
the design choices that produce them. MoR and Ouro gate the update, but Ouro supervises
every iteration and MoR needs a router; heavy-tail Poisson shares strictly and updates
unconditionally; RRT is input-independent and depth-fixed. \method{} is the only column
combining full sharing, an input-dependent update, and variable inference depth.

\begin{table}[h]
\centering
\caption{%
  \textbf{Design properties of recurrent-depth methods}, with small-scale \isoflop{}
  results for reference. Training times are wall-clock for the small configuration at
  20{,}000 steps on identical hardware, and are \emph{not} comparable against the dense
  column: the dense baseline was trained with \texttt{torch.compile} enabled, whereas
  every recurrent method including our own ran with compilation disabled, which did not
  work on our hardware. The dense column is therefore optimistic; the recurrent columns
  remain comparable to one another.
}
\label{tab:method_comparison}
\small
\setlength{\tabcolsep}{5pt}
\begin{tabular}{lcccccc}
\toprule
\textbf{Property} & \textbf{Dense} & \textbf{MoR} & \textbf{Poisson} &
\textbf{RRT} & \textbf{Ouro} & \textbf{\methodshort{}} \\
 & & {\scriptsize\citep{bae2026mixtureofrecursions}} & {\scriptsize\citep{huggin2024}}
 & {\scriptsize\citep{bae2025relaxed}} & {\scriptsize\citep{ouro}} & \textbf{(ours)} \\
\midrule
Full weight sharing            & n/a   & Yes    & Yes       & No (LoRA) & Yes   & \textbf{Yes} \\
Variable inference depth       & No    & Yes    & Yes       & No        & Yes   & \textbf{Yes} \\
Input-dependent gating         & No    & Yes    & No        & No        & Yes   & \textbf{Yes} \\
Per-step noise injection       & No    & No     & Init only & No        & No    & \textbf{Yes} \\
Per-iteration loss             & No    & No     & No        & No        & Yes   & No \\
Emergent early exit            & No    & Yes    & Yes       & No        & Yes   & \textbf{Yes} \\
\midrule
Training time (small)          & 3h18m & 12h19m & 4h16m     & 6h42m     & 8h20m & 6h18m \\
Small val.\ loss $\downarrow$  & 3.15  & 3.30   & 3.23      & 3.14      & 3.19  & \textbf{3.14} \\
\bottomrule
\end{tabular}
\end{table}

\subsection{Gate Temperature Sensitivity}

The gate temperature $\tau$ (Eq.~\ref{eq:gate}) controls the sharpness of the sigmoid.
We sweep $\tau \in \{0.5, 1.0, 2.0\}$ on the medium configuration at 5{,}000 steps.
Validation loss is 3.62, 3.60, and 3.63 respectively, indicating that the optimal gate temperature is $\tau=1.0$.

\subsection{Noise Magnitude Sensitivity}

We ablate state noise $\sigma_x \in \{0.0, 0.05, 0.1, 0.2\}$ on the medium
configuration at 5{,}000 steps.
Removing noise entirely ($\sigma_x=0.0$) increases validation loss by 0.031 nats
relative to $\sigma_x=0.1$; larger noise ($\sigma_x=0.2$) degrades by 0.018 nats.
The optimal range is $\sigma_x \in [0.05, 0.1]$; we use 0.1 throughout.

\subsection{Gate Bias and State Noise at Full Training Horizon}
\label{app:sweeps}

The sweeps above are at 5{,}000 steps. We repeated both at the full
20{,}000-step budget on the small configuration to check that the response
surface does not sharpen at convergence.
It does not.
Moving the gate bias from the default $+4$ down to $0$ costs $0.019$ nats, and
$-2$ recovers to $3.151$; the surface is shallow and non-monotonic rather than
peaked, and no setting produced a loss spike or divergence.
State noise behaves the same way: removing it costs $0.018$ nats and doubling it
costs $0.019$, a symmetric optimum rather than a value that has to be hit
precisely.
This is weaker evidence than a cross-architecture study, but it does suggest the
mechanism does not depend on the exact bias value.

\begin{table}[h]
  \centering
  \begin{minipage}[t]{0.46\linewidth}
    \centering
    \caption{Gate-bias sensitivity, small config at 20{,}000 steps.}
    \label{tab:gate_bias_sweep}
    \small
    \begin{tabular}{lc}
      \toprule
      \textbf{Gate bias} & \textbf{Val.\ loss $\downarrow$} \\
      \midrule
      $+4$ (default) & \textbf{3.141} \\
      $+2$           & 3.152 \\
      $0$            & 3.160 \\
      $-2$           & 3.151 \\
      \bottomrule
    \end{tabular}
  \end{minipage}
  \hfill
  \begin{minipage}[t]{0.46\linewidth}
    \centering
    \caption{State-noise sensitivity, small config at 20{,}000 steps.}
    \label{tab:noise_sweep}
    \small
    \begin{tabular}{lcc}
      \toprule
      $\sigma_x$ & \textbf{Val.\ loss $\downarrow$} & $\boldsymbol{\Delta}$ \\
      \midrule
      $0$              & 3.229 & $+$0.018 \\
      $0.1$ (default)  & \textbf{3.211} & --- \\
      $0.2$            & 3.230 & $+$0.019 \\
      \bottomrule
    \end{tabular}
  \end{minipage}
\end{table}

\section{Additional Qualitative Results}
\label{app:qualitative}




\subsection{Per-Token Gate Activations Across Recurrence Steps}
\label{app:gate_token}

\begin{figure}[h]
  \centering
  \includegraphics[width=\linewidth]{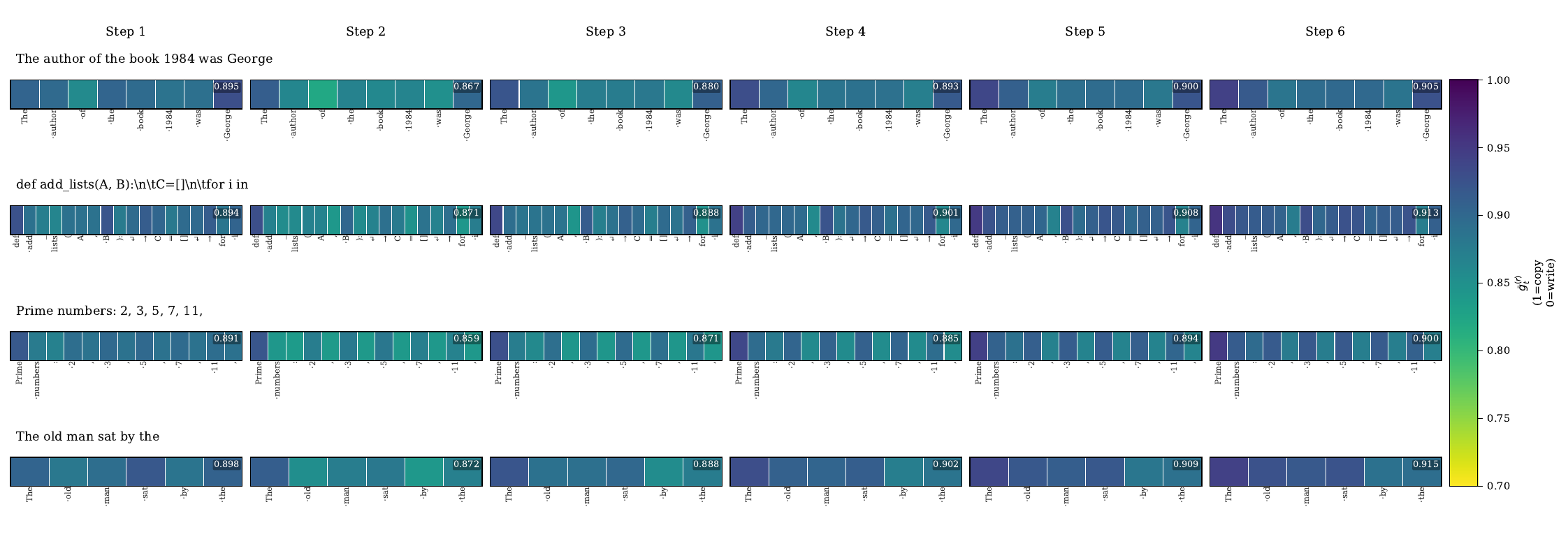}
  \caption{
    \textbf{Per-token gate activation $\bar{g}^{(r)}_t$ across recurrence steps.}
    Each row of cells corresponds to one recurrence step (columns = token positions).
    Colour encodes $\bar{g}^{(r)}_t \in [0.70, 1.00]$: bright yellow-green = write-heavy
    (gate open, block output substantially absorbed); dark purple = copy-heavy
    (hidden state passed through unchanged).
    Token labels are shown below each step column.
    The per-step mean (annotated top-right of each panel) rises from step~2 onward,
    reflecting the write-to-copy transition identified in \figref{fig:gate_behaviour}(b).
    Structurally informative tokens (content words, operators) consistently receive
    lower gate values than function words and punctuation across all steps.
  }
  \label{fig:gate_token_heatmap}
\end{figure}

\figref{fig:gate_token_heatmap} shows $\bar{g}^{(r)}_t = \mathrm{mean}_d(g^{(r)}_{t,:})$
--- the elementwise gate averaged over the model dimension --- for each token position
$t$ and recurrence step $r$, on four example prompts.
Brighter (yellow-green) cells indicate write-heavy positions where the gate
is open ($\bar{g} \approx 0.70$--$0.82$) and the shared block's proposal is
substantially absorbed; darker (purple) cells indicate copy-heavy positions
($\bar{g} \approx 0.95$--$1.00$) where the hidden state passes through largely unchanged.

Three consistent observations emerge across all four prompts.
First, \textbf{the per-step mean gate rises monotonically from step~2 onward}:
means fall from $\sim$0.87--0.89 at step~1 to a minimum around step~2
(write-heavy phase), then recover steadily toward $\sim$0.88--0.91 at step~6
(copy-heavy phase).
Second, \textbf{write-heavy positions are structurally consistent across steps}:
the same columns that light up at step~1 tend to be the same at step~6,
indicating that token identity --- not recurrence depth --- drives the gate
primary signal, with the gate progressively dimming those writes over steps
rather than shifting which tokens are written.
Third, \textbf{content words and syntactically load-bearing tokens receive lower
gate values than function words}: in ``The old man sat by the'',
\emph{man} and \emph{sat} are visibly brighter than \emph{the} and \emph{by};
in the code prompt, the operator and identifier tokens show write-heavy behaviour
while keywords like \texttt{def} and \texttt{for} are copy-heavy.
This implicit routing --- without any explicit per-token mechanism --- arises
purely from the gate conditioning on both the evolving hidden state $x^{(r)}$
and the fixed prelude anchor $h$.

\section{Broader Impact}
\label{app:impact}

Parameter-efficient language models have direct environmental and accessibility benefits:
a \method{} checkpoint at 37\% of a baseline's parameter count requires proportionally
less GPU memory at serving time, enabling deployment on hardware that cannot accommodate
the full dense model and reducing energy consumption per inference step.
Recurrent depth also provides a natural compute-quality tradeoff at inference time---the
early-exit capability of \secref{subsec:setup} allows a single trained model to serve
requests at multiple quality-FLOPs operating points without retraining.

The risks associated with \method{} are not specific to its architecture: as a capable
language model trained on web text, it inherits the standard dual-use risks of any such
system, including the potential to generate misinformation or harmful content.
These risks are addressed by standard deployment safeguards (content filtering,
output monitoring, and responsible release practices) that apply equally to all
current language models; the recurrent architecture does not introduce qualitatively
new threat vectors.

\section{Recurrence Dynamics Analysis}
\label{app:analysis}




Beyond aggregate validation loss, we probe \emph{what} the shared block
computes at each recurrence step $r \in \{1,\ldots,6\}$ using the large
\texttt{1+5$\times$6+5} checkpoint.
Diagnostics are run on a validation split with inference noise
disabled.
We report ten interlocking findings---grouped first into behavioural diagnostics
(\secref{subsec:loss_convergence}--\secref{subsec:fp}) and then into
mechanistic analyses of weights and attribution
(\secref{subsec:gate_attribution}--\secref{subsec:token_freq})---before drawing
them together in \secref{subsec:unified_summary}.

\subsection{Rapid Convergence of Loss Across Recurrence Steps}
\label{subsec:loss_convergence}

\begin{figure}[t]
  \centering
  \includegraphics[width=\linewidth]{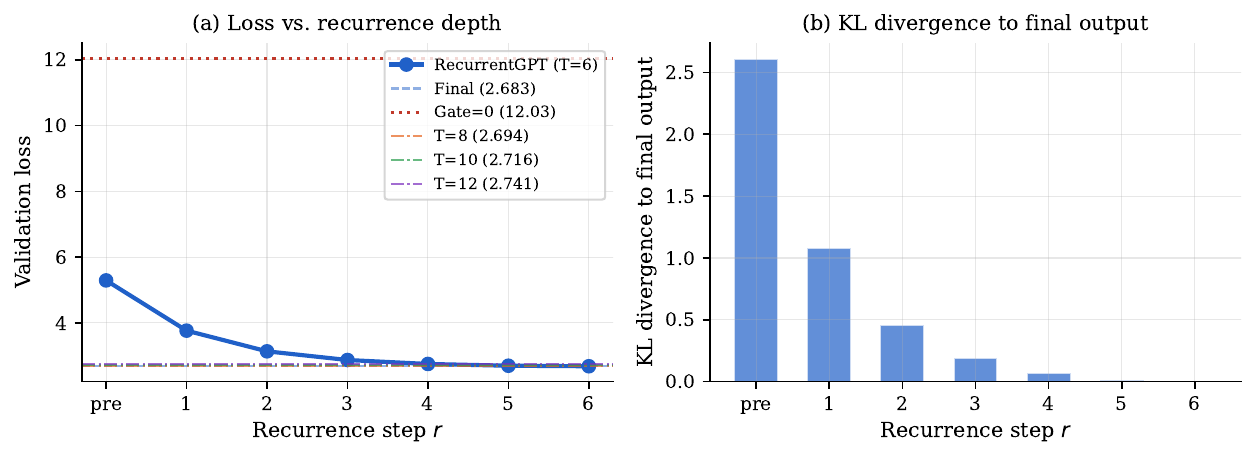}
  \caption{
    \textbf{Loss convergence across recurrence steps.}
    (a) Validation loss at each recurrence step $r$ for the trained
    \texttt{1+5$\times$6+5} model. The ``Gate=1'' baseline (loss $5.26$, green dashed) equals the prelude-only
    output, confirming that selective writing is essential: a fully open gate
    discards every block proposal, leaving the hidden state unchanged across
    all six steps.
    The ``Gate=0'' baseline (loss $12.03$) confirms that recurrence is
    essential, not a skip connection.
    Extended inference at $R\in\{8,10,12\}$ does not improve beyond $R=6$ ($2.69$, $2.72$, and $2.74$ respectively),
    indicating representational convergence.
    (b) KL divergence of each step's output distribution to the final-step
    distribution; 95\% of the residual uncertainty is resolved within four
    recurrence steps.
  }
  \label{fig:loss_convergence}
\end{figure}

\figref{fig:loss_convergence}(a) traces the per-step validation loss as the
shared block is applied iteratively.
The prelude output alone yields a loss of $5.29$; after a single application
of the shared core the loss drops to $3.77$---a reduction of $1.52$ nats (natural-log-scale bits; 1 nat $= \log_2 e \approx 1.44$ bits) in
one step.
By step~2 the loss reaches $3.14$, and the remaining four steps account for
only an additional $0.46$ nats, converging to the final value of $2.68$.
Stated differently, the first two recurrence steps account for roughly 77\%
of the total loss reduction from the prelude representation to the final output.

This front-loaded profile is confirmed by the KL-divergence trajectory in
\figref{fig:loss_convergence}(b), which measures residual uncertainty
relative to the final-step output distribution.
The KL falls from $2.61$ at the prelude to $0.46$ by step~2 and reaches
$0.014$ at step~5, indicating that the output distribution is essentially
committed after the fourth step.
As a lower bound on the gate's utility, forcing the gate to zero throughout
(the ``no-recurrence'' ablation) yields a catastrophic loss of $12.03$,
confirming that the iterative refinements are not redundant. The complementary ablation---forcing the gate to one throughout, so that every
block proposal is discarded and the hidden state is never updated---yields a loss
of $5.26$, equivalent to the prelude-only output ($5.26$ at step~$r=0$).
This confirms that recurrence with a fully open gate is vacuous: without selective
writing, six recurrence steps reduce to a single prelude pass.
Together, Gate=0 ($12.03$) and Gate=1 ($5.26$) bracket the trained model ($2.68$):
the learned gate neither skips recurrence nor blindly overwrites state, but
acquires a fine-grained elementwise balance between retention and update.

We also evaluate extending the recurrence depth at inference time beyond the
$R=6$ steps seen during training.
Applying the trained shared block for $R \in \{8, 10, 12\}$ steps
marginally \emph{degrades} performance ($2.69$, $2.72$, $2.74$ respectively),
suggesting the model has converged representationally by step~6 and that
additional steps introduce noise rather than refinement.

\begin{takeaway}
  77\% of the loss reduction from prelude to final output occurs within the first
  two recurrence steps; the remaining four steps provide targeted corrections.
  The model's output distribution is essentially committed by step~4.
\end{takeaway}
\subsection{Gate Behaviour: Write-Heavy Early, Copy-Heavy Late}
\label{subsec:gate}

\begin{figure}[t]
  \centering
  \includegraphics[width=\linewidth]{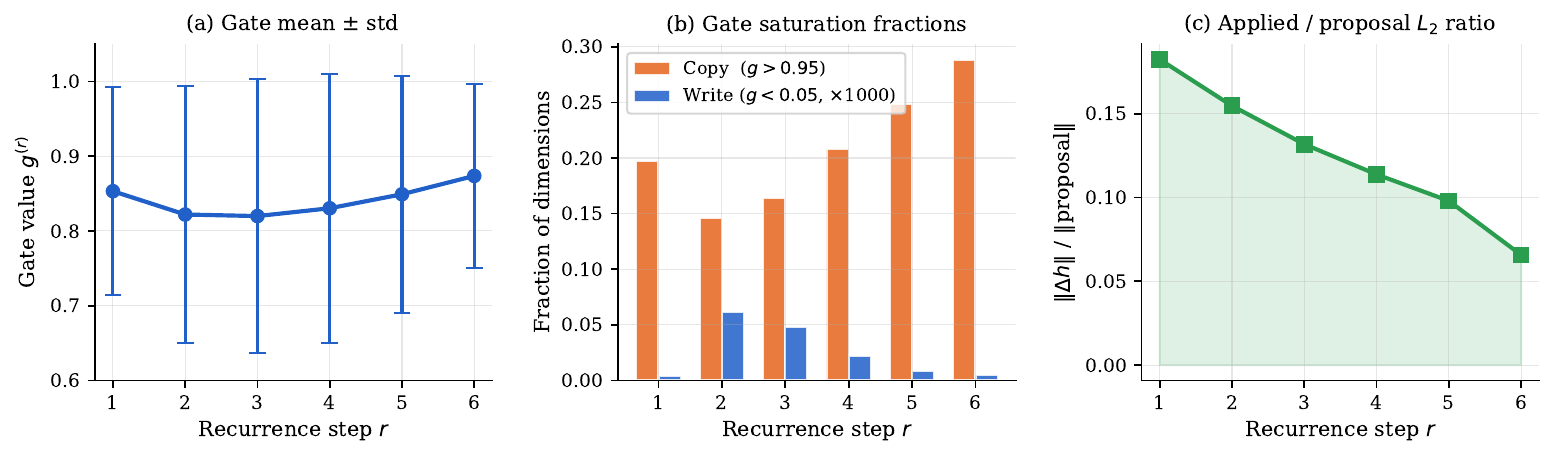}
  \caption{
    \textbf{Gate statistics across recurrence steps.}
    (a) Gate mean and standard deviation.
    (b) Fraction of copy-saturated ($g > 0.95$) and write-saturated
    ($g < 0.05$, scaled $\times 1000$) dimensions per step.
    Copy-saturation grows steadily while write-saturation stays negligible.
    (c) Effective gate openness (applied/proposal $L_2$ ratio), declining
    monotonically---early steps write substantially, late steps refine conservatively.
  }
  \label{fig:gate_behaviour}
\end{figure}

The elementwise gate $\gate^{(r)} \in [0,1]^{S \times d}$ controls how much
of the shared block's proposal is written into the residual stream at each
step.
\figref{fig:gate_behaviour}(a) plots the per-step mean and standard deviation.
The gate is most open---and most variable---in the middle steps (steps 2--3,
mean $\approx 0.82$, std $\approx 0.18$), corresponding to the highest
information-writing regime, and tightens progressively toward step~6
(mean $0.87$, std $0.12$).

\figref{fig:gate_behaviour}(b) quantifies this transition via the fraction of
dimensions where the gate is near-saturated.
Copy-saturated dimensions ($g > 0.95$) grow monotonically from 19.7\% at
step~1 to 28.8\% at step~6, while write-saturated dimensions ($g < 0.05$)
remain negligibly rare ($<10^{-4}$) throughout.
This asymmetry---abundant copying, scarce full-overwriting---indicates the
model prefers selective blending over clean state replacement.

The effective gate openness (\figref{fig:gate_behaviour}(c)), measured as the ratio
\begin{equation}
  \rho^{(r)} = \frac{\|\hstate^{(r)} - \hstate^{(r-1)}\|_2}{\|\mathbf{o}^{(r)} - \hstate^{(r-1)}\|_2}
             = \frac{\|(1 - \gate^{(r)}) \odot (\mathbf{o}^{(r)} - \hstate^{(r-1)})\|_2}{\|\mathbf{o}^{(r)} - \hstate^{(r-1)}\|_2},
  \label{eq:gate_openness}
\end{equation}
where the numerator is the norm of the \emph{applied update} (actual change to the hidden state)
and the denominator is the norm of the \emph{proposal} (the full update the block would write
if the gate were fully open), decreases
monotonically from $0.182$ at step~1 to $0.066$ at step~6.
Taken together, the gate transitions the model from a \emph{write-heavy}
regime in early steps---where large fractions of the proposal are incorporated
to rapidly restructure the representation---to a \emph{copy-heavy} regime in
later steps, where the representation is largely preserved and only small
targeted corrections are applied.

\begin{takeaway}
  The gate transitions from a write-heavy regime (steps 1--2, mean $\approx 0.82$)
  to a copy-heavy regime (steps 3--6, mean $\approx 0.87$), with effective gate
  openness declining monotonically from $0.18$ to $0.07$.
\end{takeaway}
\subsection{Update Dynamics: Diminishing Magnitude, Increasing Alignment}
\label{subsec:update}

\begin{figure}[t]
  \centering
  \includegraphics[width=\linewidth]{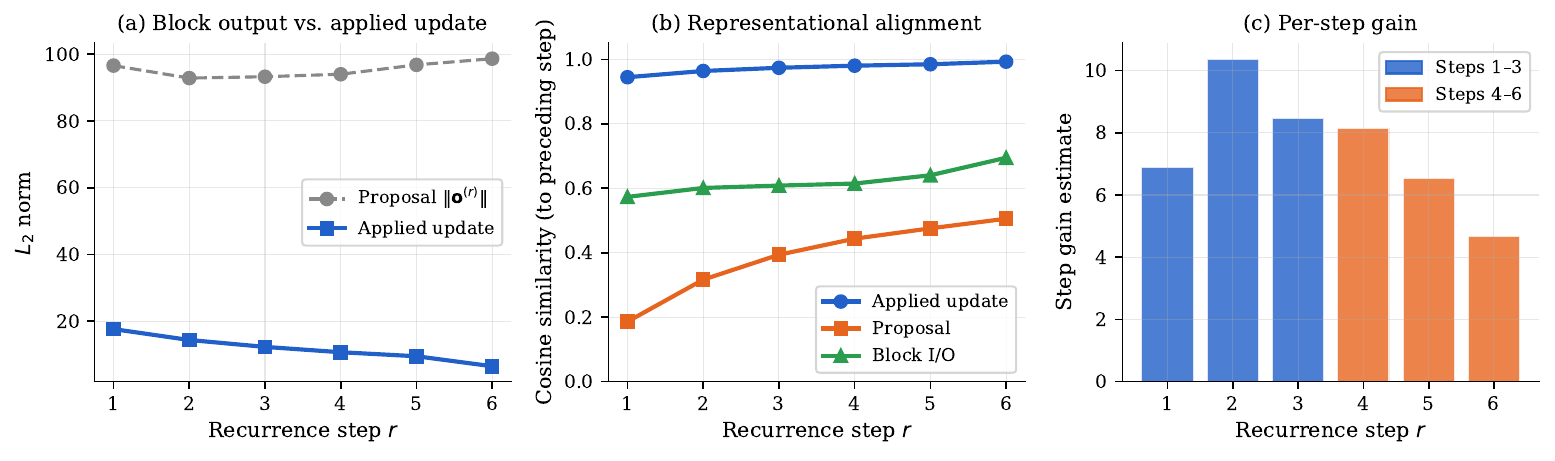}
  \caption{
    \textbf{Update dynamics across recurrence steps.}
    (a) $L_2$ norms of the block's raw proposal (near-constant, $\approx 94$)
    and the gate-applied update (declining $17.6 \to 6.5$); the divergence is
    caused entirely by gate closure.
    (b) Cosine similarity between successive representations.
    (c) Local step gain estimate, peaking at step~2 and declining thereafter.
  }
  \label{fig:update_dynamics}
\end{figure}

\figref{fig:update_dynamics}(a) separates two quantities at each step: the
$L_2$ norm of the shared block's raw proposal $\mathbf{o}^{(r)}$, and the
$L_2$ norm of the gate-applied update that actually enters the residual stream.
The proposal norms are nearly constant across all six steps ($93$--$99$),
possibly indicating that the shared block does not itself ``know'' it is being applied
repeatedly---its output magnitude is stationary.
The applied update, by contrast, declines from $17.6$ at step~1 to $6.5$ at
step~6, driven entirely by the gate becoming more closed.

\figref{fig:update_dynamics}(b) shows cosine similarity between successive
representations, $\text{cos\_sim}(\hstate^{r}, \hstate^{r-1})=\frac{\hstate^{r} \cdot \hstate^{r-1}}{\| \hstate^{r} \| \| \hstate^{r-1} \|}$.
The proposal cosine similarity grows from $0.19$ at step~1 to $0.51$ at step~6,
indicating that the block and the hidden state increasingly agree on direction
as the representation stabilises.
The applied update cosine similarity is already high at step~1 ($0.94$) and
approaches $0.99$ by step~6, confirming that late-step corrections are
directionally consistent refinements.

The per-step gain estimate (\figref{fig:update_dynamics}(c)) peaks at
step~2 ($10.4$), declining to $4.7$ at step~6.
This concave gain profile, in which the most productive computation
occurs before representational alignment is achieved, is consistent with
the loss convergence plots in \secref{subsec:loss_convergence}.

\subsection{Representational Geometry: Two-Phase Structure}
\label{subsec:cka}

We use Centered Kernel Alignment~\citep[CKA;][]{kornblith2019similarity} to
compare hidden-state representations across recurrence steps and against the
corresponding layer representations of the GPT-2 Large dense baseline.

\begin{figure}[t]
  \centering
  \includegraphics[width=\linewidth]{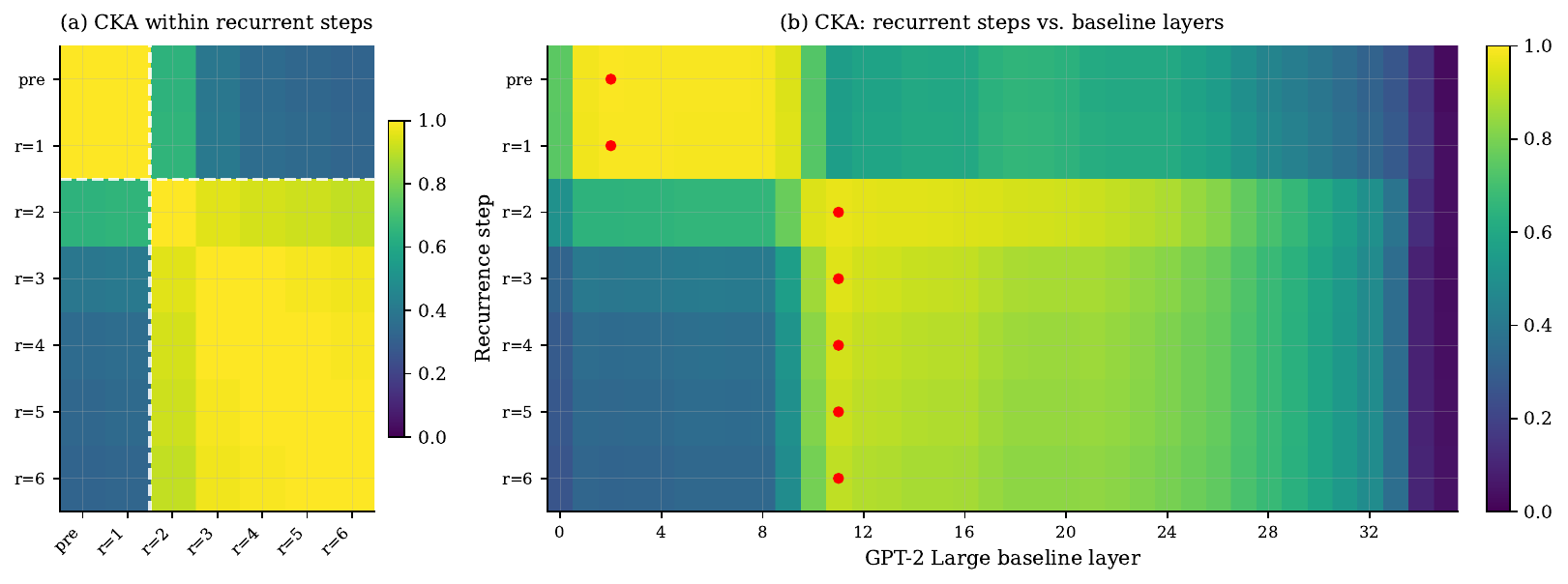}
  \caption{
    \textbf{CKA representational geometry.}
    (a) Within-model CKA; the sharp discontinuity between steps 1--2 and
    steps 3--6 (dashed lines) reveals a two-phase structure.
    (b) Cross-model CKA against GPT-2 Large; early steps map to shallow
    baseline layers, all later steps peak at layer $\approx 11$.
  }
  \label{fig:cka}
\end{figure}

\paragraph{Within-model CKA} (\figref{fig:cka}(a)).
The $7 \times 7$ CKA matrix reveals a two-phase structure.
Representations at step~0 (prelude) and step~1 are nearly identical
(CKA $= 0.996$), while both are sharply dissimilar from all subsequent steps
(CKA $\approx 0.33$--$0.65$ with steps $\geq 2$).
Steps 2 through 6 form a tightly cohesive cluster (pairwise CKA $\geq 0.907$),
within which similarity decreases only gradually with step distance.
This suggests the model undergoes a \emph{representational phase transition}
between steps 1 and 2.

\paragraph{Cross-model CKA} (\figref{fig:cka}(b)).
Prelude and step~1 representations align maximally with the shallow layers of
the baseline (peak at layer 2, CKA $= 0.993$), while steps~2 through~6 all
peak at baseline layer~11 (CKA $= 0.97$--$0.90$ declining).
No recurrent step aligns strongly with the deep layers of the baseline
(layers $>20$, CKA $< 0.50$), confirming that recurrent depth compresses
the computational path but does not replicate the later representational
hierarchy of the dense model.

\begin{takeaway}
  A representational phase transition occurs between steps 1 and 2: early steps
  resemble shallow baseline layers, while steps 2--6 all align with mid-depth
  baseline layer~11, compressing the computational path without replicating
  the deep-layer hierarchy.
\end{takeaway}

\subsection{Differential Routing by Token Difficulty}
\label{subsec:difficulty}

\begin{figure}[t]
  \centering
  \includegraphics[width=\linewidth]{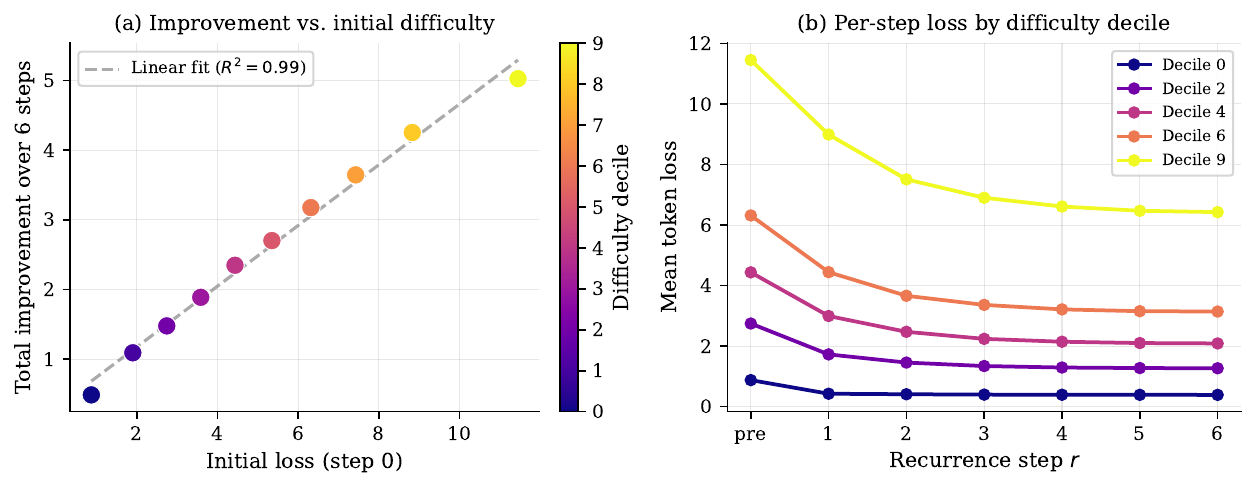}
  \caption{
    \textbf{Token difficulty routing.}
    (a) Total improvement over six steps vs.\ initial difficulty decile
    ($R^2 = 0.998$); recurrence implicitly allocates more work to uncertain tokens.
    (b) Per-step loss curves for selected deciles.
  }
  \label{fig:token_difficulty}
\end{figure}

To test whether the gate allocates computation to where it is most needed,
we sort tokens by their loss after the prelude (step~0) into ten deciles and
track the total loss improvement accumulated over six recurrence steps.
\figref{fig:token_difficulty}(a) shows a near-linear relationship between
initial token difficulty and total improvement ($R^2 = 0.998$): the easiest
decile gains only $0.49$ nats, while the hardest decile gains $5.02$ nats---a
ten-fold difference.
The gate difficulty correlation---whether tokens with higher loss also receive
lower (more open) gate values---is mildly negative across all steps
($-0.07$ to $-0.04$), consistent with the gate allowing slightly more writing
for uncertain tokens.

\subsection{Fixed-Point Convergence and Effective Rank}
\label{subsec:fp}

\begin{figure}[t]
  \centering
  \includegraphics[width=\linewidth]{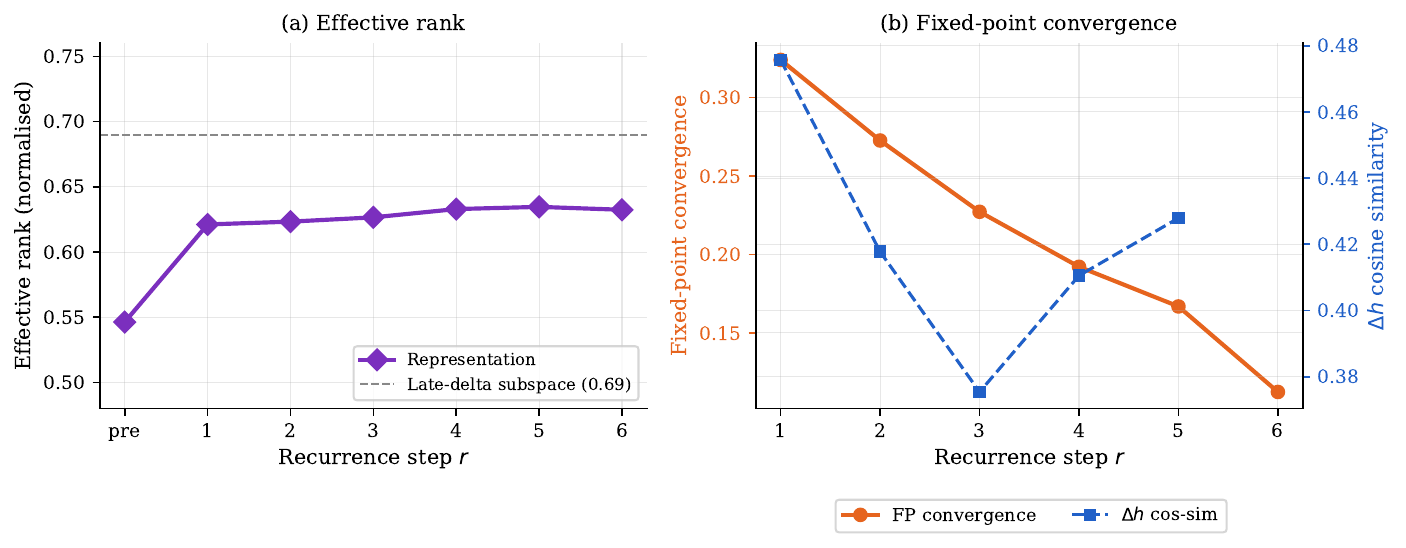}
  \caption{
    \textbf{Effective rank and fixed-point convergence.}
    (a) Effective rank grows with depth; the late-delta subspace rank (dashed)
    is higher than the representation itself.
    (b) Fixed-point convergence (orange) declines monotonically; $\Delta h$
    cosine similarity (blue) increases---both consistent with convergent refinement.
  }
  \label{fig:rank_convergence}
\end{figure}

\figref{fig:rank_convergence}(a) tracks the effective rank of the hidden-state
tensor at each step.
The rank increases from $0.546$ at the prelude to $0.635$ by step~3, after
which it plateaus, indicating that iterative updates progressively spread
information across more hidden dimensions.
The effective rank of the \emph{delta subspace}---the directions along which
the hidden state changes in later steps---is $0.69$, higher than the rank of
the representation itself, suggesting that late-step updates refine a diverse
set of features.

\figref{fig:rank_convergence}(b) plots the normalised fixed-point convergence
metric, defined as
\begin{equation}
  \delta_{\text{fp}}^{(r)} = \frac{\|\hstate^{(r)} - \hstate^{(r-1)}\|_F}{\|\hstate^{(r-1)}\|_F},
  \label{eq:fp_convergence}
\end{equation}
the Frobenius-norm magnitude of the update relative to the current representation,
which declines monotonically from $0.324$ at step~1 to $0.112$ at
step~6, confirming that the recurrent dynamics are contractive.

\tabref{tab:anchor_ablation} reports a sanity check via anchor ablations---swapping
the fixed prelude output $\hstate^{(\mathrm{pre})}$ fed at each recurrence step for alternative
anchors while keeping the gate and blocks identical.
Crucially, the ``previous hidden state'' row uses $\hstate^{(r-1)}$ as anchor (i.e.\ the
anchor drifts with the recurrence rather than being held fixed), while the learned gate
remains fully active and adaptive.
The $0.70$-nat gap between this condition ($3.38$) and the trained model ($2.68$) therefore
quantifies the value of providing a \emph{stable, fixed reference point} at every step,
not merely the value of the gate.

\begin{table}[h]
  \centering
  \small
  \caption{%
    Anchor ablation: effect of replacing $\hstate^{(\mathrm{pre})}$ with alternative
    anchors at each recurrence step.
    The ``prev.\ hidden state'' anchor drifts at each step ($\hstate^{(r-1)}$) while
    the learned adaptive gate remains active throughout.
    The $0.70$-nat gap between that condition and the trained model isolates the
    value of holding the prelude output \emph{fixed} as a stable context reference.
  }
  \label{tab:anchor_ablation}
  \begin{tabular}{lcc}
    \toprule
    \textbf{Anchor type} & \textbf{Val.\ loss (nats)} & $\Delta$ vs.\ trained \\
    \midrule
    Zeros                       & 8.08 & +5.40 \\
    Input embedding             & 3.73 & +1.05 \\
    Prev.\ hidden state $\hstate^{(r-1)}$ & 3.38 & +0.70 \\
    \midrule
    \textbf{Trained model} (fixed $\hstate^{(\mathrm{pre})}$) & \textbf{2.68} & --- \\
    \bottomrule
  \end{tabular}
\end{table}

\subsection{Gate Attribution: Current State Gradually Supersedes the Anchor}
\label{subsec:gate_attribution}

\begin{figure}[t]
  \centering
  \includegraphics[width=\linewidth]{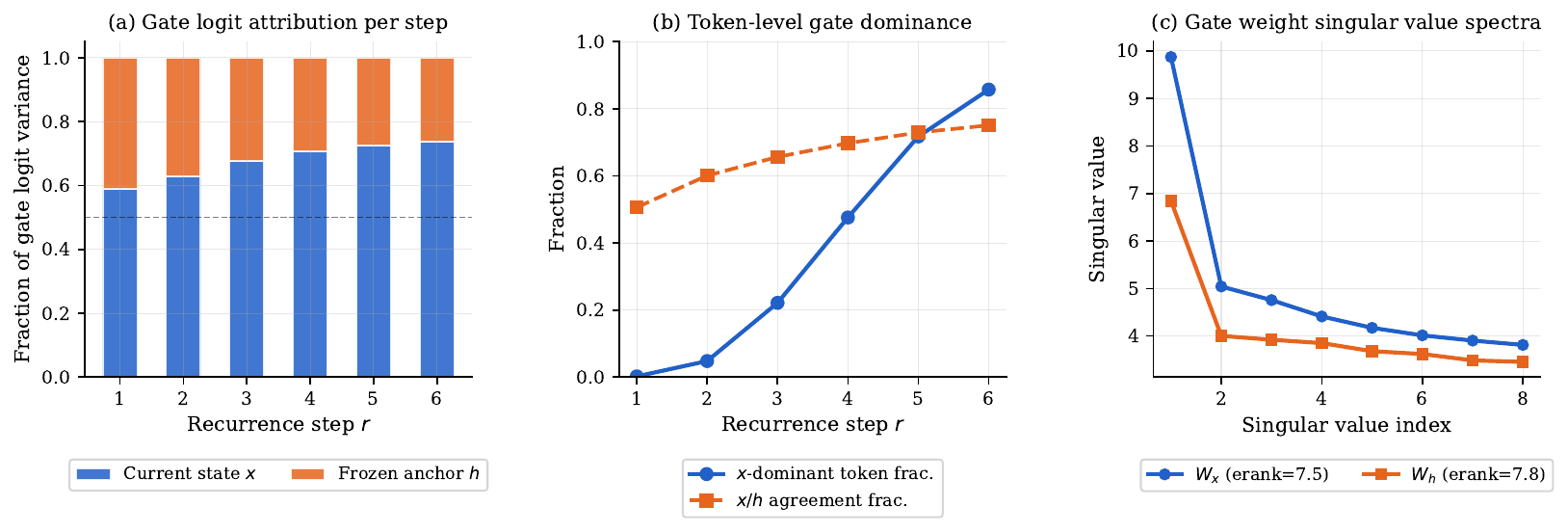}
  \caption{
    \textbf{Gate attribution across recurrence steps.}
    (a) Fraction of gate logit variance from current state $x$ (blue) vs.\
    frozen anchor $h$ (orange); the crossover from rough parity at step~1 to
    74\%/26\% at step~6 is explained by growing divergence between $x^{(r)}$ and $h$.
    (b) Token-level $x$-dominance and $x/h$ sign-agreement fractions.
    (c) Gate weight singular value spectra; the 44\% gap in leading singular
    values ($9.87$ vs $6.85$) is the structural origin of the state's eventual
    dominance.
  }
  \label{fig:gate_attribution}
\end{figure}

The elementwise gate (Eq.~\ref{eq:gate}) reads from two sources simultaneously:
$\hat{\mathbf{x}}^{(r)} \triangleq \mathrm{LN}(\hstate^{(r-1)})$,
the LayerNorm-normalised current hidden state,\footnote{We write
$\hat{\mathbf{x}}^{(r)}$ rather than $x$ to avoid confusion with token IDs
$(x_1,\ldots,x_S)$ introduced in the Preliminaries.}
and $\hat{\mathbf{h}} \triangleq \mathrm{LN}(\hstate^{(\mathrm{pre})})$,
the normalised prelude anchor.
The gate MLP $f_\gate$ takes their concatenation as input; we write its
first linear layer weight as $W_\gate = [W_x \mid W_h] \in \reals^{d_{\text{gate}} \times 2d}$,
where $W_x \in \reals^{d_{\text{gate}} \times d}$ acts on $\hat{\mathbf{x}}^{(r)}$
and $W_h \in \reals^{d_{\text{gate}} \times d}$ acts on $\hat{\mathbf{h}}$.
To quantify which source drives the gate at each step, we measure how much variance in the
actual per-token gate logit each half explains.
As seen in \figref{fig:gate_attribution}(a), at step~1 the split is nearly even: 59\% current state, 41\% anchor.
By step~6 this shifts substantially---74\% current state, 26\% anchor.

\figref{fig:gate_attribution}(b) shows the same trend at the token level:
the fraction of positions where the state half produces a larger absolute
gate logit than the anchor grows from a near-zero 0.2\% at step~1 to 86\%
at step~6.
The answer to \emph{why} is present in the weights before any forward
pass: we find in \figref{fig:gate_attribution}(c) that the leading singular value of $W_x$ is 9.87 versus 6.85 for $W_h$, giving the state half structurally greater sensitivity.
At step~1 the hidden state is close to the prelude output, so the gap is small;
as $x^{(r)}$ diverges from $h$ across steps, this structural asymmetry is
amplified into the monotone attribution shift in \figref{fig:gate_attribution}(a).

\begin{takeaway}
  By step~6, the gate reads 74\% of its signal from the current hidden state
  and only 26\% from the fixed prelude anchor.
  This shift is structurally predetermined: $W_x$ has a 44\% larger leading
  singular value than $W_h$, priming the gate to amplify state-driven signals
  as $\hstate^{(r)}$ departs from $\hstate^{(\mathrm{pre})}$.
\end{takeaway}

\subsection{Projection Weight Analysis: State and Anchor as Complements}
\label{subsec:proj_weights}

Before the shared blocks run, the recurrent projection, $W_{\text{proj}}$, maps the concatenated
$[x^{(r)},\, h]$ into the $d$-dimensional input the transformer sees.
A natural question is whether this projection simply averages the two sources
or treats them as carrying distinct information.
The answer is unambiguous: the mean row-wise cosine similarity between the
$W_x$ and $W_h$ halves is $\mathbf{-0.189}$, with only $0.16\%$ of the 1024
rows aligned above a cosine of $0.5$.

\begin{figure}[t]
  \centering
  \includegraphics[width=\linewidth]{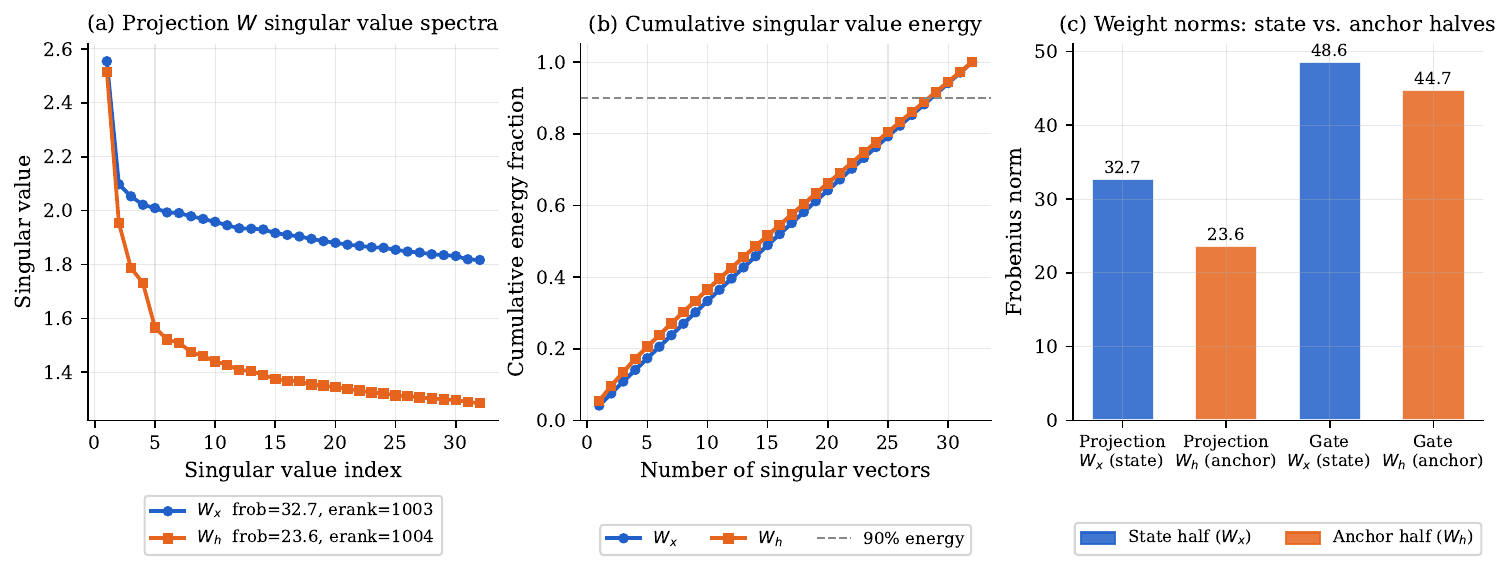}
  \caption{
    \textbf{Recurrent projection weight analysis.}
    (a) Singular value spectra of $W_x$ and $W_h$; both are nearly full-rank.
    (b) Cumulative energy; both halves reach 90\% at similar dimensionalities.
    (c) Frobenius norms of all four weight halves; the state half ($W_x$)
    consistently outweighs the anchor half ($W_h$) in both modules.
  }
  \label{fig:proj_weights}
\end{figure}

Both halves are nearly full-rank (effective ranks 1002.9 and 1004.3,
Frobenius norms 32.7 and 23.6), with flat, slowly-decaying singular value
spectra (\figref{fig:proj_weights}(a)--(b)).
The negative mean row cosine means the projection reads the
\emph{difference} between current state and anchor rather than their average;
we call this the \emph{contrastive projection} property.
Directions large in both $x^{(r)}$ and $h$---stable features unchanged since
the prelude---partially cancel; directions large only in $x^{(r)}$ are
amplified.
This provides a direct mechanistic explanation for why the fixed prelude is
helpful: it supplies the reference needed to compute what has changed.

\subsection{Attention Entropy: Stable Head Specialisations, Gradual Sharpening}
\label{subsec:attn_entropy}

\begin{figure}[t]
  \centering
  \includegraphics[width=\linewidth]{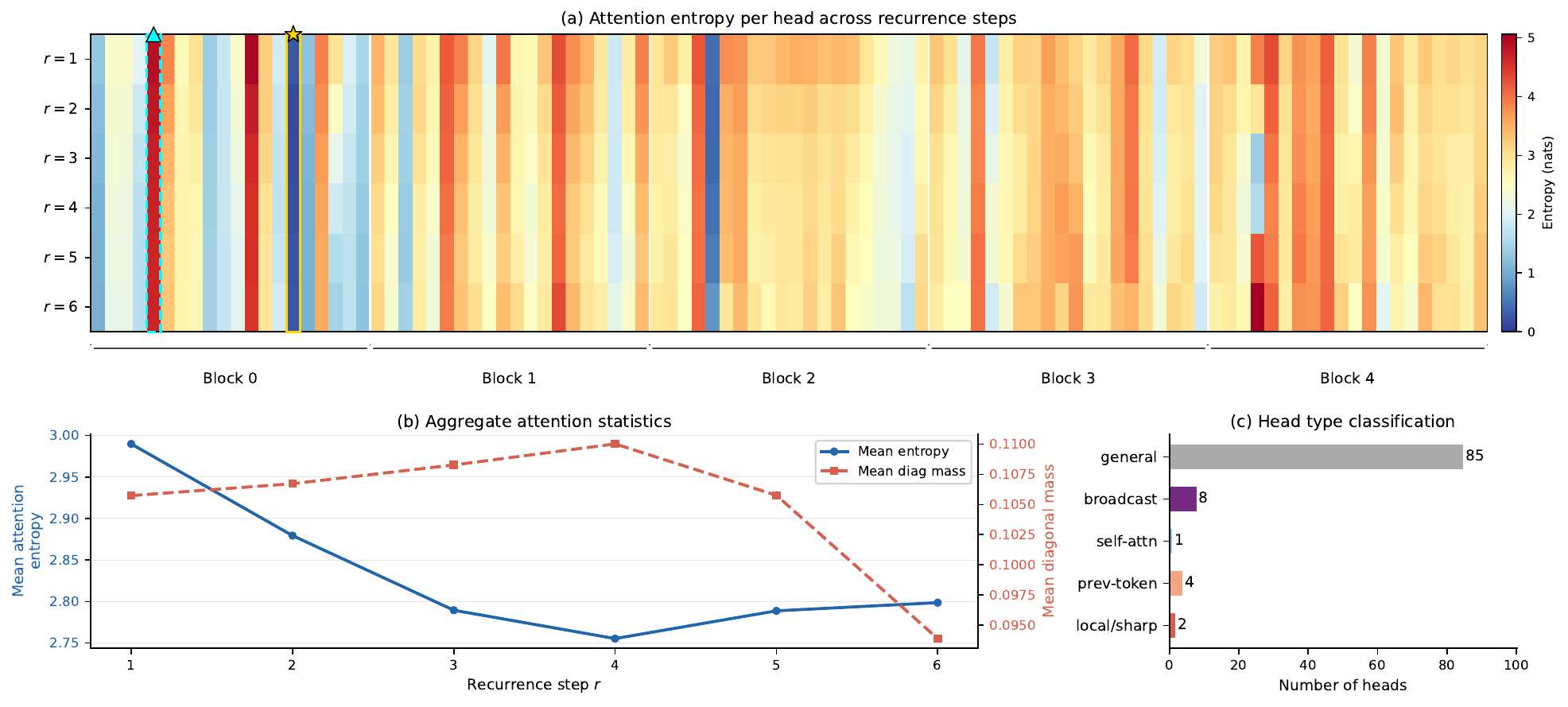}
  \caption{
    \textbf{Attention entropy across recurrence steps.}
    (a) Full $6 \times 100$ entropy heatmap; white lines separate blocks.
    $\star$: Block~0 Head~14 (previous-token specialist, entropy 0.30).
    $\triangle$: Block~0 Head~4 (broadcast head, entropy 4.87).
    Head roles are highly stable.
    (b) Mean entropy (blue) and diagonal mass (orange) per step; non-monotone
    entropy profile mirrors the two-phase CKA structure.
    (c) Head type classification across all 100 heads.
  }
  \label{fig:attn_entropy}
\end{figure}

Because identical weights govern every recurrence step, the shared block
cannot self-modulate its attention patterns---any change must arise purely
from changes in $\hstate^{(r)}$.
In \figref{fig:attn_entropy}(a), we capture the full $[B, n_\text{head}, S, S]$ attention tensors at each step
(where $B$ is batch size and $n_\text{head}$ the number of attention heads)
across all five shared blocks (100 heads total) and measure per-head Shannon
entropy and diagonal mass.

\paragraph{Aggregate trend.}
Mean attention entropy declines from 3.00 nats at step~1 to 2.77 at step~4,
then \emph{rebounds} to 2.82 at steps 5--6 (\figref{fig:attn_entropy}(b)).
This non-monotone profile mirrors the two-phase CKA structure from
\secref{subsec:cka}: steps~1--4 are the refinement phase (lower entropy,
more peaked patterns), while the rebound at steps~5--6 aligns with the final
convergence plateau where small corrective updates scan broadly.
Mean diagonal mass peaks at step~4 (0.108) then drops to 0.092 at step~6.

\paragraph{Head role stability.}
The full $6 \times 100$ entropy heatmap (\figref{fig:attn_entropy}(a)) makes
the dominant pattern immediately visible: individual heads retain their
character across all six steps.
Of 100 heads, 2 are local/sharp (entropy $< 1.0$), 3 are previous-token
(prev-mass $> 0.35$), 2 are self-attention (diagonal mass $> 0.35$), 7 are
broad broadcast (entropy $> 4.0$), and 86 are general.
Here, \emph{prev-mass} denotes the fraction of total attention weight
placed on the immediately preceding token position (i.e.\ diagonal offset $-1$
of the attention matrix), averaged over the sequence.
A head with prev-mass $> 0.35$ predominantly attends to the token immediately
before each query position.
The clearest specialist---Block~0, Head~14---has entropy $0.298$ and
prev-mass $0.878$ at step~$r=1$,

\begin{takeaway}
  Head specialisations are locked in by weight and do not adapt to recurrence
  depth. What changes across steps is only the magnitude of each head's
  contribution, modulated by the gated update to the hidden state.
\end{takeaway}

\subsection{Recurrence Preferentially Serves Two Token Regimes}
\label{subsec:token_freq}

\begin{figure}[t]
  \centering
  \includegraphics[width=\linewidth]{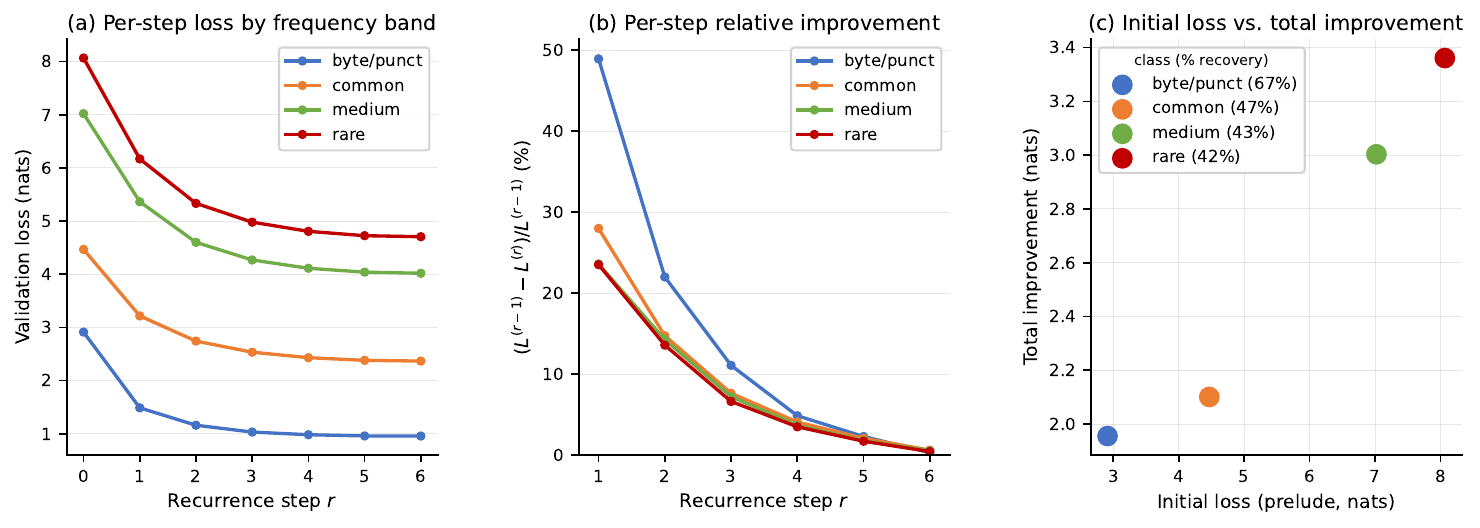}
  \caption{
    \textbf{Token frequency class breakdown.}
    (a) Per-step loss by BPE frequency band; rare tokens maintain persistently
    higher loss throughout.
    (b) Per-step relative improvement $(L^{(r-1)}-L^{(r)})/L^{(r-1)}$ per frequency class;
    the first recurrence step alone resolves 47\% of byte/punctuation loss
    vs.\ 23\% for rare tokens---easy tokens converge fastest, hard tokens
    accumulate improvements gradually.
    (c) Initial loss vs.\ total improvement; the two axes are near-orthogonal,
    consistent with dual completion/narrowing roles.
  }
  \label{fig:token_categories}
\end{figure}

We split tokens into four BPE frequency bands: byte/punctuation (IDs 0--255,
$n=5{,}763$), common (IDs 256--2{,}000, $n=15{,}789$), medium (IDs 2{,}001--10{,}000,
$n=6{,}778$), and rare (IDs $>10{,}000$, $n=4{,}438$), where $n$ refers to the total occurrences of corresponding token id ranges in the validation set.
\figref{fig:token_categories}(a) shows that all four follow the familiar
front-loaded loss-reduction profile, but at very different absolute scales:
byte/punctuation tokens improve by just 1.92 nats while rare tokens improve
by 3.39 nats.

\figref{fig:token_categories}(b) plots the per-step relative improvement
$\Delta_r^{\mathrm{rel}} = (L^{(r-1)} - L^{(r)}) / L^{(r-1)}$, i.e.\
the fraction of the \emph{preceding step's} loss resolved at each step.
At step~$r=1$, byte/punctuation tokens already resolve 49\% of their previous
loss, compared to 23--28\% for the other bands---recurrence is
disproportionately productive for easy tokens first.
Gains diminish uniformly beyond step~3 ($< 4$\% for all bands),
consistent with the fixed-point convergence of \secref{subsec:fp}.
Cumulatively over all six steps, byte/punctuation achieves 67\% recovery of its
initial loss versus 42\% for rare tokens.
We call these two roles \emph{completion} (sharpening already-likely token predictions)
and \emph{lexical narrowing} (progressively concentrating probability mass for hard tokens).
Both operate simultaneously in the same forward pass, possible because the
elementwise gate allocates computation per-position rather than globally---no
explicit routing mechanism is required.

\subsection{Unified Summary}
\label{subsec:unified_summary}

The ten analyses above converge on a coherent computational and mechanistic
account of how \method{}'s shared block achieves its parameter efficiency.

Behaviourally, the model operates in two regimes separated by a
representational phase transition between steps~1 and~2.
In the \emph{early regime} (steps 1--2), the gate is relatively open ($\approx
0.82$ mean, large update norms, high proposal cosine entropy), the hidden
state undergoes a rapid restructuring that resolves $\sim$77\% of predictive
uncertainty, and CKA confirms a sharp departure from the prelude representation.
In the \emph{late regime} (steps 3--6), the gate progressively closes, updates
shrink and align, the representation approaches a fixed point, and the attention
patterns partially rebroadcast---consistent with targeted correction rather
than coarse restructuring.

Mechanistically, four findings explain \emph{why} this behaviour emerges from
the learned weights.
The \emph{gate weight asymmetry} ($W_x$ leading singular values 44\% larger than $W_h$)
primes the gate to respond more strongly to changes in the hidden state than
to the anchor, an advantage that is dynamically amplified as $x^{(r)}$ departs
from $h$ across steps.
The \emph{contrastive projection} (mean row cosine $-0.19$) ensures the shared
block receives an input that emphasises what has changed since the prelude,
giving each step access to a difference signal rather than a raw state.
The \emph{head specialisations} are fixed by weight---the same 14 previous-token
heads, 8 broadcast heads, and 11 self-attention heads appear at every step---so
the block applies the same functional program each time; the gate, not the
heads, decides how much of each step's output to incorporate.
Finally, the \emph{token frequency breakdown} reveals that a single forward pass
simultaneously performs completion for common tokens and lexical narrowing for
rare ones, made possible because the elementwise gate implicitly routes
computation per-position without any explicit routing mechanism.

\end{document}